\documentclass[journal]{IEEEtran}
\usepackage[T1,OT1]{fontenc}
\usepackage{cite}
\usepackage{microtype}
\usepackage{booktabs}
\usepackage{multirow}
\usepackage{array}
\usepackage{tabularx,listings,textcomp}
\usepackage{amsmath,amssymb}
\usepackage{graphicx}
\usepackage{url}
\usepackage{algorithm}
\usepackage{algpseudocode}
\newcolumntype{Y}{>{\raggedright\arraybackslash}X}
\newcommand{\clip}{\operatorname{clip}}
\title{CurEvo: Curriculum-Guided Self-Evolution for Video Understanding}
\author{Guiyi Zeng, Junqing Yu, Yi-Ping Phoebe Chen, Chen Xu, Wei Yang, and Zikai Song%
\thanks{Guiyi Zeng, Junqing Yu, Wei Yang, and Zikai Song are with the Huazhong University of Science and Technology, Wuhan, China.}
\thanks{Yi-Ping Phoebe Chen is with La Trobe University, Melbourne, Australia.}
\thanks{Chen Xu is with the Beijing Institute of Computer Technology and Applications, Beijing, China.}
\thanks{Corresponding author: Zikai Song (e-mail: skyesong@hust.edu.cn).}}
\begin{document}
\maketitle

\begin{abstract}
Recent self-evolution frameworks for video understanding have demonstrated the potential for autonomous improvement from unlabeled videos. However, existing approaches often suffer from weakly controlled optimization and capability progression because they lack structured guidance for generating, evaluating, and reusing supervision throughout the iterative learning process. We present \textbf{CurEvo}, a curriculum-guided self-evolution framework that organizes this loop as a competence-aware progression. \textbf{Curriculum-Guided Self-Evolution} maintains a competence-aware curriculum state and cumulative training pool across iterations. Within this loop, \textbf{Multi-Dimensional Question Generation} supplies complementary supervision, from directly verifiable structural questions for \textbf{perception} to type-conditioned questions for \textbf{recognition} and \textbf{understanding}. \textbf{Type-Adaptive Evaluation} separately assesses question quality and answer reliability with criteria matched to each type's evidence; its scores determine which pairs enter the pool and how the next curriculum state is updated. Across seven backbones and four VideoQA benchmarks, CurEvo consistently improves benchmark accuracy and benchmark semantic score. For Qwen2.5-VL(7B), accuracy rises from 67.84\% to 69.91\% on ActivityNet-QA and from 76.94\% to 79.61\% on NExT-QA. CurEvo thus establishes self-generated supervision as a stable route to autonomous video understanding without task-specific human QA annotations during self-evolution.
\end{abstract}


\begin{figure}[t]
    \centering
    \includegraphics[width=\linewidth]{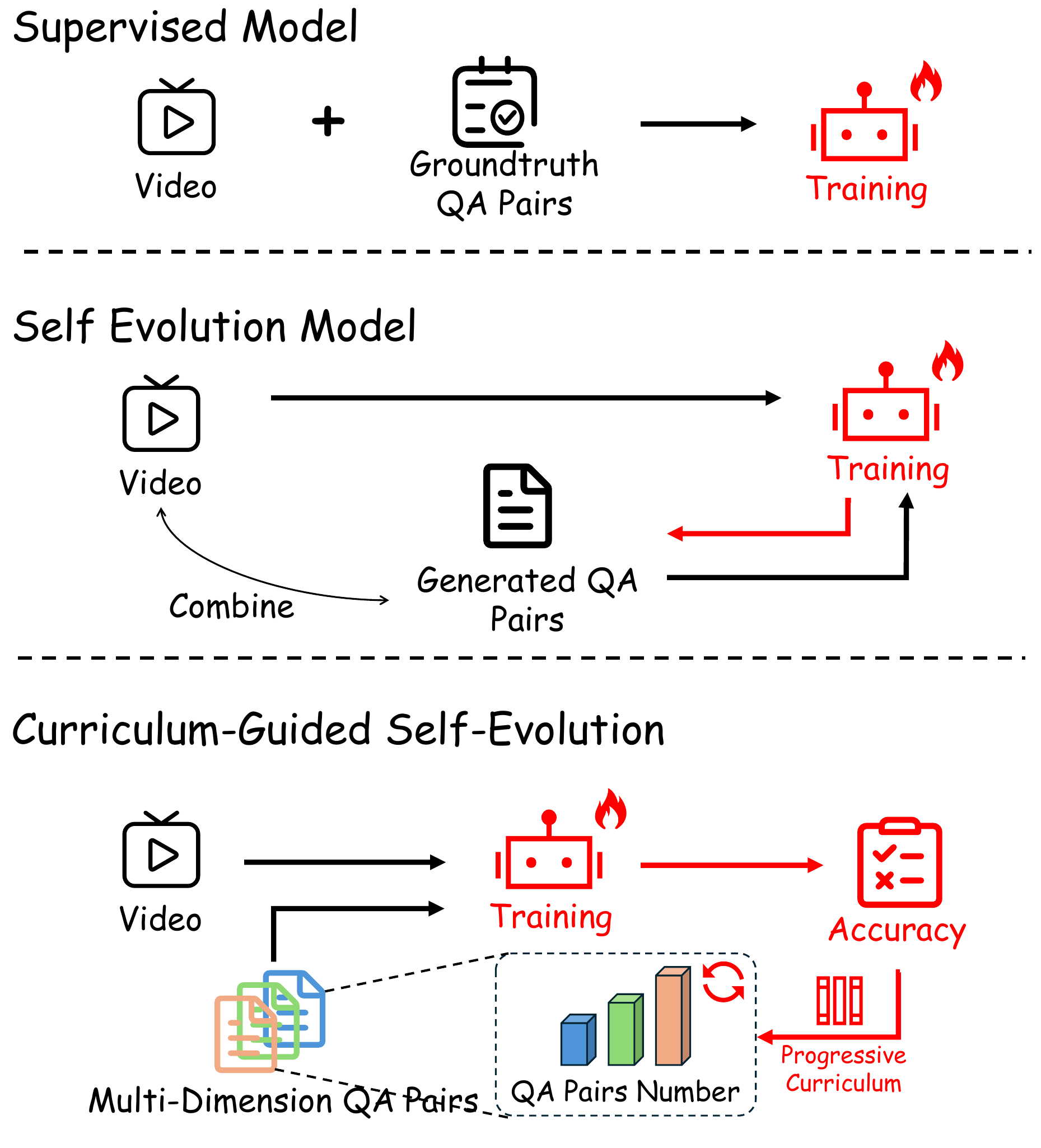}
    \caption{Comparison of supervised, conventional self-evolving, and CurEvo training. Supervised VideoQA relies on a fixed human-annotated distribution, while conventional self-evolution repeatedly retrains on self-generated pairs without an explicit capability trajectory. CurEvo combines self-generated question--answer pairs with type-adaptive selection and a curriculum state that organizes supervision and model updates over successive iterations.}
    \label{fig:teaser}
\end{figure}

\section{Introduction}
\label{sec:intro}
Video question answering (VideoQA) requires a model to interpret visual content, track temporal events, and produce an answer in natural language or a task-specific format. Recent video large language models (Video-LLMs) combine visual encoders with language models to support this cross-modal temporal reasoning~\cite{jeshmol2024video, nguyen2024video, madan2024foundation, wang2025videotree, lu2026question, huang2026video}, yet their performance remains closely tied to the supervision available during training. As illustrated in Fig.~\ref{fig:teaser}, fixed human-annotated datasets provide stable targets but cover limited scenes and question types. Models can consequently learn visual--language alignment while remaining tied to annotated-data coverage. The same supervision may also encourage models to exploit superficial visual--text correlations instead of learning temporal and causal relations~\cite{xiao2023contrastive, gao2023mist, tang2025video, tang2025video1, kang2026causality}. After training on this fixed distribution, no explicit mechanism remains for adapting to new reasoning demands or revisiting capabilities that need further supervision.

Self-evolution provides a way to extend video understanding beyond a fixed annotation pipeline. Recent approaches generate question--answer pairs, evaluate their outputs, and retrain the same model in an iterative loop~\cite{yang2021just, qiu2025step, zohar2024video, chen2022debiased, shi2025enhancing}. Deriving supervision from unlabeled videos reduces dependence on additional labels and supports repeated updates. However, generated supervision varies in quality, semantic depth, and difficulty across iterations~\cite{xia2023learning, chen2022debiased, wang2022debiased, chen2025self}. Without an explicit learning trajectory, unreliable answers may enter training, easier question types may dominate the pool, and the model has no principled signal for deciding where additional supervision is most useful. Effective self-evolution therefore requires supervision that is trustworthy at the example level and aligned with capability-level needs.

We address this gap with \textbf{CurEvo}, a curriculum-guided self-evolution framework for VideoQA. CurEvo organizes self-evolution as a competence-conditioned supervision loop: it constructs complementary questions, verifies each generated pair with evidence matched to its capability, and uses the resulting feedback to determine subsequent allocation while retaining reliable records. This linkage couples sample-level reliability with capability-level progression, thereby structuring self-generated data as a reusable supervision source with directed capability coverage~\cite{karim2023c, li2024answering}.

CurEvo realizes this loop through three coordinated stages in which each stage supplies the state or supervision required by the next. \textbf{Curriculum-Guided Self-Evolution} maintains the curriculum state and cumulative training pool, allowing accepted supervision to persist across iterations while the state determines the next sampling and update step. Given this state, \textbf{Multi-Dimensional Question Generation} constructs complementary questions across perception, recognition, and understanding, using directly verifiable structural questions for perception and type-conditioned questions for recognition and understanding. \textbf{Type-Adaptive Evaluation} then assesses each question--answer pair with evidence-matched question and answer gates; its scores determine pool membership and training weights, and update the curriculum state for the following iteration.

Across seven backbone models and four VideoQA benchmarks, CurEvo improves both benchmark accuracy and benchmark semantic score. The gains cover short-video grounding, long-range temporal understanding, and causal reasoning, and the ablations show that the three supervision dimensions provide complementary signals. On ActivityNet-QA, Qwen2.5-VL(7B) improves from 67.84\% to 69.91\%; the complete configuration also outperforms the external self-evolution baselines in the controlled ablation. Together, these results support structured quality selection and curriculum-guided reuse as an effective way to support targeted self-evolution across model families and capability levels.

\begin{figure*}[!t]
    \centering
    \includegraphics[width=\linewidth]{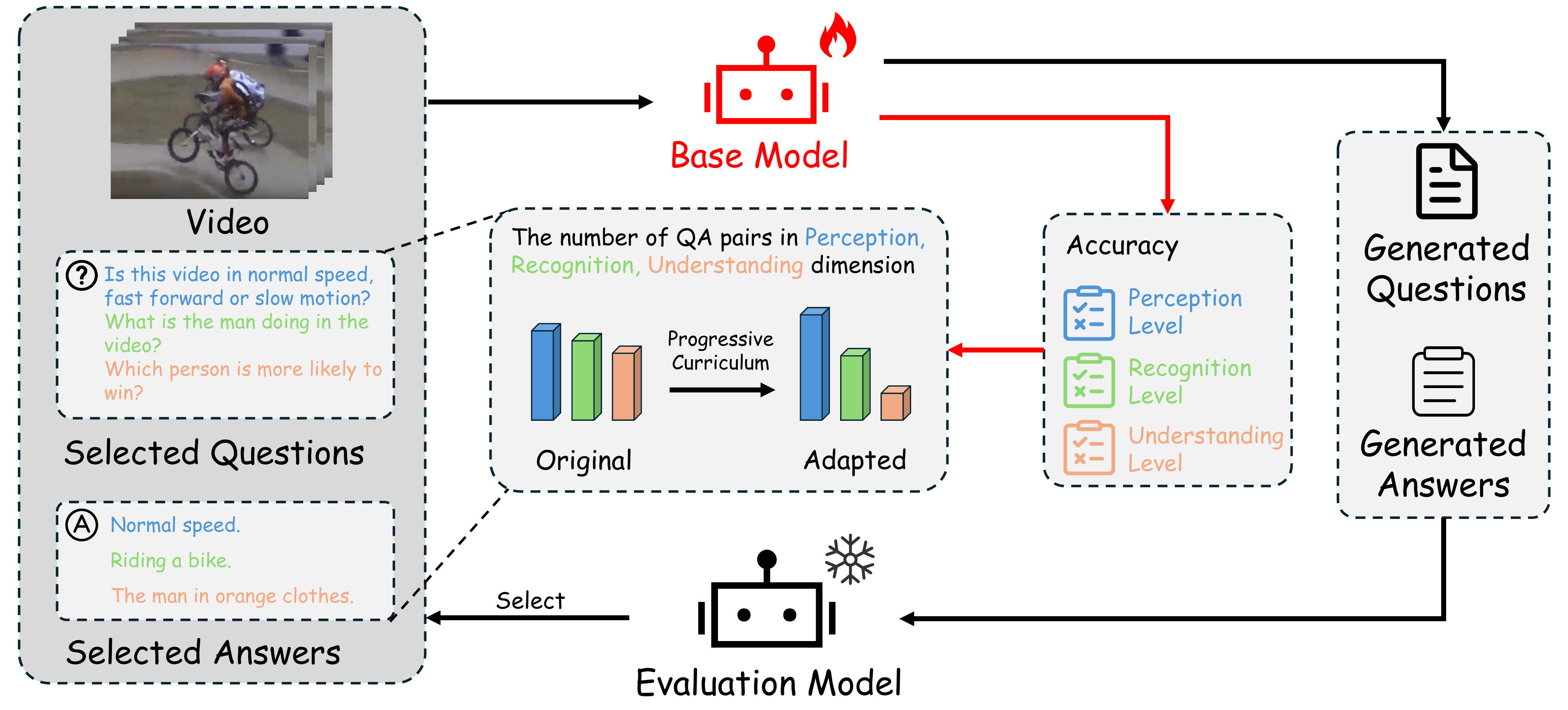}
    \caption{CurEvo organizes video self-evolution as a quality-selective curriculum loop. Dimension-aware ratios assign videos before candidate generation, questions are screened before answer generation, answers are verified with type-appropriate evidence, and accepted pairs are accumulated in a weighted pool for model updates while current-round scores update the next curriculum state.}
    \label{fig:pipeline}
\end{figure*}

\section{Related Work}
\label{sec:related_works}
CurEvo is positioned at the intersection of Video-LLMs, self-evolution, and curriculum learning. Existing work largely advances these directions independently, whereas CurEvo connects them through a competence-aware loop that controls the reliability and capability coverage of self-generated VideoQA data~\cite{song2026socialintelligence}.

\subsection{Video Large Language Models}
Video-LLMs extend vision-language modeling to dynamic scenes by combining visual representations with temporal modeling and language-based reasoning~\cite{sun2019videobert, zhang2023video, maaz2024video, guo2021universal, zhang2019frame}. Early video-language models such as VideoBERT, CLIP4Clip, Frozen-in-Time, and Flamingo established cross-modal transfer and zero-shot video understanding~\cite{sun2019videobert, luo2022clip4clip, bain2021frozen, alayrac2022flamingo}. More recent systems unify video encoders and autoregressive decoders to support open-ended dialogue and reasoning over longer temporal contexts~\cite{yang2023self, li2023videochat}. This progression makes Video-LLMs a flexible interface for video evidence and language generation and provides the base-model setting for CurEvo to construct and reuse supervision~\cite{ye2026omnitrendcontentcontextmodelingscalable}.

Most Video-LLMs rely on static annotated or text-aligned data, with limited control over generated supervision's quality and capability coverage. CurEvo organizes unlabeled-video supervision across grounding, recognition, and reasoning as competence changes.

\subsection{Self-Evolution for Video-LLMs}
Self-evolution improves a model by repeatedly constructing and reusing supervision from unlabeled data. Classical self-training and bootstrapping introduced pseudo-label refinement, while later self-supervised methods used representation consistency to obtain learning signals without manual labels~\cite{scudder1965probability, yarowsky1995unsupervised, abney2002bootstrapping, he2020momentum, grill2020bootstrap}. Multimodal systems have extended this idea with feedback loops between generation, evaluation, and retraining for Video-LLMs~\cite{madaan2023self, kedrick2023multifaceted, asai2024self, song2025modularized, wang2025videorft, li2026admitting, wu2026scl}. These approaches reduce the need for additional human labels, but they commonly leave the reliability and capability composition of generated supervision weakly controlled. CurEvo focuses on organizing this loop so that evidence-specific selection and competence feedback jointly determine which supervision is retained and which capability receives additional coverage.

A recurring limitation is that self-evolution treats accepted pairs as a flat pool without preserving an explicit competence trajectory. CurEvo addresses this gap with type-aware evaluation and cumulative pool management, connecting evidence-specific reliability to a capability curriculum.

\subsection{Curriculum Learning}
Curriculum learning organizes training examples so that a model gradually acquires more complex skills~\cite{bengio2009curriculum}. Self-Paced Learning and Self-Paced Curriculum Learning established adaptive pacing, and later work explored teacher models and reinforcement learning for automated curricula~\cite{kumar2010self, jiang2015self, narvekar2020curriculum, portelas2020automatic, wang2021survey, akl2024task}. These approaches show that supervision order and composition affect optimization and generalization~\cite{zhang-etal-2026-logical}. They also provide a useful perspective for VideoQA, where perceptual grounding, recognition, and reasoning require different kinds of evidence and different degrees of abstraction~\cite{xia2022video}.

Most curricula assume labeled data or a fixed difficulty signal. CurEvo embeds three-dimensional question generation, type-adaptive filtering, and competence feedback in the self-evolution loop.

\section{Challenges and Motivation}
\label{sec:challenge}
Self-evolving VideoQA transforms unlabeled videos into supervision and feeds evaluation outcomes into model updates. Sustained capability growth therefore depends on resolving three challenges within this loop: the \textbf{lack of a competence-aware evolution path} across successive iterations, \textbf{unstructured capability coverage} in generated supervision, and \textbf{unreliable supervision verification} across heterogeneous question--answer pairs.

\paragraph{Challenges in Self-Evolution}

\textbf{Lack of a Competence-Aware Evolution Path} Existing loops repeat generation, evaluation, and retraining without an explicit state connecting current competence to subsequent sampling, weighting, and reuse. As relative competence gaps change, these loops offer no principled mechanism for reallocating learning emphasis, while reliable supervision from earlier rounds may play no persistent role in later updates. \textbf{Curriculum-Guided Self-Evolution} addresses this challenge with a curriculum state and a cumulative weighted training pool: the state redistributes sampling emphasis as competence gaps evolve, and the pool preserves accepted supervision across iterations.

\textbf{Unstructured Capability Coverage} Self-generated supervision must cover different kinds of video understanding rather than form an undifferentiated question--answer pool. We denote the candidates recorded at iteration $t$ by
\begin{equation}
\mathcal{C}_t=\mathcal{C}_{\mathrm{perc},t}\cup\mathcal{C}_{\mathrm{recog},t}\cup\mathcal{C}_{\mathrm{under},t},
\label{eq:candidate_decomposition}
\end{equation}
where the three subsets correspond to perception, recognition, and understanding. If the dimensions are not constructed deliberately, easier or more frequent question types can dominate and leave the model with incomplete capability coverage. \textbf{Multi-Dimensional Question Generation} addresses this challenge by supplying complementary question types across the three dimensions, using directly verifiable structural questions for perception and type-conditioned questions for recognition and understanding.

\textbf{Unreliable Supervision Verification} Generated questions and answers vary in reliability, while the evidence needed to verify them depends on their capability dimension. A perception question may have a directly checkable target, whereas an understanding question may require semantic assessment of an open-ended answer; a single global criterion cannot account for both. \textbf{Type-Adaptive Evaluation} addresses this challenge by separating question and answer assessment and applying gates matched to the evidence available for each question type. Its scores determine whether a generated pair enters the training pool and how strongly it contributes to optimization.

\paragraph{Design Objective}
Together, the three challenges define a single objective: construct complementary candidates, retain only evidence-supported question--answer pairs, and direct their training influence according to the model's current competence state. \textbf{Multi-Dimensional Question Generation} establishes the candidate space $\mathcal{C}_t$; \textbf{Type-Adaptive Evaluation} checks each candidate with question and answer criteria matched to its evidence and yields the accepted set $\mathcal{A}_t$; and \textbf{Curriculum-Guided Self-Evolution} merges accepted records into the cumulative pool, reweights the retained supervision, and updates $s_{t+1}$. This ordering links coverage, reliability, and progression in one loop: current-round feedback redirects allocation while accepted records remain available for later updates.

\section{Method}
\label{sec:method}
Figure~\ref{fig:pipeline} presents CurEvo as a recurrent process that converts unlabeled videos into reliable supervision. At iteration $t$, the curriculum state $s_t$ specifies the target capability dimensions and question types, the generator constructs complementary candidates from these assignments, and the QA-pool evaluator determines which candidates qualify for the next model update. \textbf{Curriculum-Guided Self-Evolution} supplies the state and preserves accepted supervision; \textbf{Multi-Dimensional Question Generation} converts each assignment into complementary perception, recognition, and understanding candidates; and \textbf{Type-Adaptive Evaluation} verifies those candidates with evidence-matched criteria for pool admission, training influence, and feedback for the following state. Algorithm~\ref{alg:curevo} summarizes this recurrent procedure; the implementation details appear in the supplement.

\begin{algorithm}[t]
\caption{Curriculum-guided self-evolution}
\label{alg:curevo}
\begin{algorithmic}[1]
    \State Initialize model parameters $\Theta^{(0)}$, curriculum state $s_0$, and pool $\mathcal{P}_0$
\For{$t=0,\ldots,T-1$}
    \State Sample videos and assign dimensions and question types using $s_t$
    \State Generate and record candidate set $\mathcal{C}_t$; the QA-pool evaluator scores questions and applies $Q_i\geq\tau^q_{k_i}$
    \State Generate base answers for the retained questions
    \State Apply answer gates $A_i\geq\tau^a_{d_i}$ and $G_i=1$ to non-empty answers; obtain accepted set $\mathcal{A}_t$
    \State Merge the accepted set with the historical pool
    \State Compute current raw priorities and retain $\mathcal{P}_{t+1}\gets\operatorname{TopM}(\mathcal{P}_t\cup\mathcal{A}_t;\widetilde\omega^{(t)})$
    \State Normalize retained weights and update $\Theta^{(t+1)}$ on $\mathcal{P}_{t+1}$
    \State Update thresholds and $s_{t+1}$ from all current-round scored records
    \State Save $\mathcal{P}_{t+1}$, thresholds, and $s_{t+1}$
\EndFor
\State \Return $\Theta^{(T)}$
\end{algorithmic}
\end{algorithm}

\subsection{Curriculum-Guided Self-Evolution}
The central role of \textbf{Curriculum-Guided Self-Evolution} is to represent each iteration as a state transition rather than an isolated retraining step. Current competence determines the target capability coverage, accepted records remain available for later updates, and the resulting feedback changes the next state. Sequential filtering, competence-dependent allocation, and cumulative weighting implement this transition.

\paragraph{Self-Evolution Cycle}
At iteration $t$, the current state $s_t$ assigns each video to target capability dimensions and question types. The base model $f_{\Theta^{(t)}}$ generates recognition and understanding questions, whereas controlled transformations construct perception questions with deterministic targets. A frozen QA-pool evaluator $g_{\Phi}$ assesses generated content, and deterministic matching verifies perception answers. This order is deliberate: weak questions are removed before answer generation, so the training pool is built only from pairs that pass both question and answer checks. For candidate $i$, let $q_i,a_i$ be its question and answer, $Q_i,A_i$ their scores, and $G_i\in\{0,1\}$ a gate for format, refusal, and pseudo-label validity. The question survivors are

\begin{equation}
\mathcal{C}_t^q=\left\{i\in\mathcal{C}_t:\ Q_i\geq\tau^q_{k_i}\right\},
\label{eq:question_survivor}
\end{equation}
Only these survivors are answered, and the second gate then produces the accepted set
\begin{equation}
\mathcal{A}_t=\left\{i\in\mathcal{C}_t^q:\ a_i\neq\varnothing,\ A_i\geq\tau^a_{d_i},\ G_i=1\right\}.
\label{eq:accepted_answer_set}
\end{equation}
Here $k_i$ and $d_i$ denote the question type and capability dimension, with thresholds $\tau^q_{k_i}$ and $\tau^a_{d_i}$. Thus, a pair influences optimization only after both evidence-appropriate gates pass.

Accepted records augment rather than replace earlier supervision. Let $\mathcal{U}_t=\mathcal{P}_t\cup\mathcal{A}_t$ be the merged pool before truncation. We retain
\begin{equation}
\mathcal{P}_{t+1}=\operatorname{TopM}\left(\mathcal{U}_t;\widetilde\omega^{(t)}\right),
\label{eq:pool_update}
\end{equation}
where $M$ is the pool limit and $\widetilde\omega^{(t)}$ the current raw priorities. Recomputing priorities before truncation allows historical records to be reconsidered under the current curriculum. The retained pool updates the model, while current-round feedback produces $s_{t+1}$ and determines the capability coverage targeted in the next iteration. The cumulative pool preserves supervision that passes reliability checks, while the frozen QA-pool evaluator provides a stable reference as generated answers evolve. The updated state then determines subsequent generation, completing the competence-aware evolution path introduced in Section~\ref{sec:challenge}.

\paragraph{Competence-Dependent Allocation}
Current-round answer-correctness feedback provides the competence signal for allocating supervision in the next iteration. CurEvo converts each dimension's correctness gap into a relative allocation rather than a fixed schedule. Let $r_d^{(t)}$ be the sampling ratio and $c_d^{(t)}$ the mean correctness over non-empty answers in dimension $d$. Perception uses pseudo-label matching; recognition and understanding use QA-pool evaluator feedback. If no record is available, $c_d^{(t)}=0$. With competence gap $g_d^{(t)}=1-c_d^{(t)}$, we update
\begin{equation}
    \begin{aligned}
    u_d^{(t)}&=r_d^{(t)}\left[1+\lambda_d g_d^{(t)}\right],\\
    \bar r_d^{(t+1)}&=\frac{u_d^{(t)}}{\sum_{d'}u_{d'}^{(t)}},\\
    r_d^{(t+1)}&=\frac{\max\!\left(\rho,\bar r_d^{(t+1)}\right)}
    {\sum_{d'}\max\!\left(\rho,\bar r_{d'}^{(t+1)}\right)}.
    \end{aligned}
\label{eq:ratio_update}
\end{equation}
Here $\lambda_d$ controls how strongly the competence gap changes a dimension's emphasis, and $\rho$ prevents any dimension from disappearing; both sums cover the three dimensions. The normalization allows a dimension to gain or lose emphasis while preserving nonzero coverage. Because the ratios act before question generation, they express requested coverage rather than the final accepted-pair composition, which is also shaped by generation, deduplication, and quality gates. The next allocation is therefore recomputed from the model's current answers, keeping the recurrent loop tied to observable supervision feedback; the reported coefficient values are given in the supplementary material.

\paragraph{Cumulative-Pool Weighting}
Once new accepted pairs are merged with the historical pool, CurEvo determines which records remain and how strongly they contribute to optimization. Equal treatment would ignore changing capability needs and sample reliability, so each priority combines the record's dimension, answer reliability, and informativeness. Here, $U_i$ summarizes normalized answer length, lexical diversity, and specificity, $d_i$ is the capability dimension, and $N_{t+1}=|\mathcal{P}_{t+1}|$:
\begin{equation}
\begin{aligned}
\widetilde\omega_i^{(t)}&=r_{d_i}^{(t)}\max(0,A_i)U_i,\qquad i\in\mathcal{U}_t,\\
\omega_i^{(t)}&=\frac{\widetilde\omega_i^{(t)}}
  {\sum_{j\in\mathcal{P}_{t+1}}\widetilde\omega_j^{(t)}},\qquad i\in\mathcal{P}_{t+1},\\
\bar\omega_i^{(t)}&=N_{t+1}\omega_i^{(t)}.
\end{aligned}
\label{eq:sample_weight}
\end{equation}
Top-$M$ retention uses the raw priorities before they are normalized over the retained pool. Multiplication by $N_{t+1}$ gives mean-one effective weights, with a uniform fallback if all priorities vanish. The dimension term encodes the curriculum balance, the answer term favors reliable records, and $U_i$ distinguishes informative records within a dimension~\cite{li2024answering}. For batch $B$, let $T_i$ be the unmasked positions and $y_{i,u}$ the target token:
\begin{equation}
\ell_i=\frac{1}{|T_i|}\sum_{u\in T_i}\operatorname{CE}(y_{i,u},\hat y_{i,u}),\qquad
\mathcal{L}_B=\frac{1}{|B|}\sum_{i\in B}\bar\omega_i^{(t)}\ell_i.
\label{eq:final_loss}
\end{equation}
Here $\hat y_{i,u}$ is the token distribution. Perception uses deterministic targets; recognition and understanding use base-model-generated answers that pass answer evaluation. Weighting after token averaging prevents answer length from controlling an example's contribution~\cite{mao2023cross}. The weighted pool updates the model, while current-round answer feedback returns to the allocation step, so each iteration starts from both the learner's competence and the supervision already retained. The next subsection describes how assigned videos yield candidate questions.

\begin{figure}[!t]
    \centering
    \includegraphics[width=\linewidth]{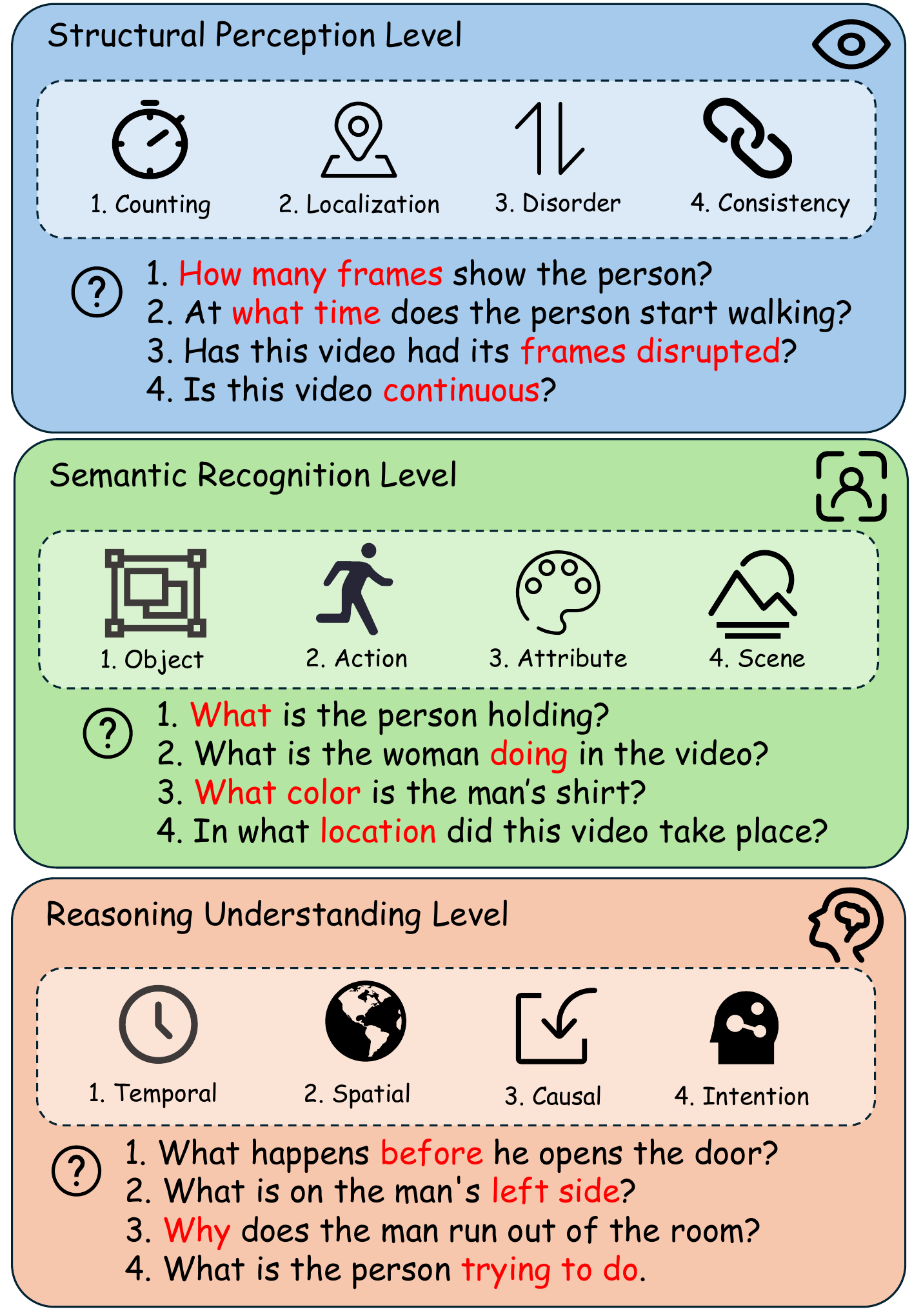}
    \caption{Question generation across perception, recognition, and understanding. The three dimensions combine directly verifiable cues with increasingly compositional video-language questions.}
    \label{fig:question}
\end{figure}

\subsection{Multi-Dimensional Question Generation}
To prevent the requested curriculum from concentrating on a single question pattern, \textbf{Multi-Dimensional Question Generation} maps each assignment to complementary candidate supervision, with perception capturing directly verifiable structure, recognition describing visible content, and understanding connecting events and relations. Figure~\ref{fig:question} illustrates the resulting progression. Each candidate retains its question type and capability dimension so that the subsequent evaluation can apply evidence-matched checks.

\paragraph{Perception Level}
Controlled temporal or structural transformations construct questions about order, count, localization, or consistency~\cite{hu2025sf2t, qian2023locate}. Their transformation-derived targets make correctness directly checkable and provide deterministic pseudo labels for curriculum allocation.

\paragraph{Recognition Level}
Recognition questions cover visible objects, actions, attributes, and scenes. Question-type guidance enables the base model to generate video-specific content while preserving the type information needed for evaluation~\cite{su2021end, zhou2025scene}; the generated question is admitted as supervision only after it passes the evaluation stage.

\paragraph{Understanding Level}
Understanding questions extend guided generation to temporal order, spatial relations, causality, and intent. Because they require evidence across fragments and allow semantically equivalent answers~\cite{zhang-etal-2026-semantic}, they undergo semantic verification before pool insertion. Together, the three levels form the structured candidate set $\mathcal{C}_t$ and preserve the distinction between directly checkable perception evidence and semantically judged recognition or understanding evidence; normalization, deduplication, and choice handling are detailed in the supplementary material. The resulting candidates are then passed to Type-Adaptive Evaluation for evidence-matched screening.

\subsection{Type-Adaptive Evaluation}
\textbf{Type-Adaptive Evaluation} requires both the question and its answer to provide dependable supervision. It first screens question quality and then evaluates surviving answers with dimension-matched evidence. Question and answer scores control acceptance, answer scores contribute to weighting, and current-round statistics update the thresholds and competence state.

\paragraph{Question Assessment}
Question assessment precedes answer generation because malformed or weakly grounded prompts cannot provide useful training targets. The first gate combines structural checks with QA-pool evaluator semantics:
\begin{equation}
    Q_i=\alpha H_q(q_i)+(1-\alpha)S_q(q_i),
\label{eq:question_score}
\end{equation}
Here $H_q=(h_{\mathrm{len}}+h_{\mathrm{form}}+h_{\mathrm{shape}}+h_{\mathrm{clear}})/4$ covers length, punctuation, question form, and leakage; $S_q$ averages semantic alignment, visual grounding, and answerability; and $\alpha$ is a fixed mixing coefficient~\cite{huang2019watch, xiao2024can, zou2024language}. Missing semantic fields trigger the heuristic fallback. A question passes when $Q_i\geq\tau^q_{k_i}$.

\begin{figure}[t]
    \centering
    \includegraphics[width=\linewidth]{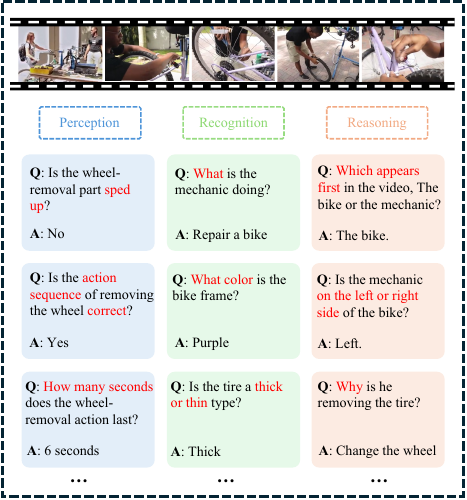}
    \caption{Examples of CurEvo-generated question--answer pairs. The examples illustrate how the same video can provide complementary perception, recognition, and understanding supervision while supporting semantic recognition and higher-level reasoning.}
    \label{fig:example}
\end{figure}

\paragraph{Answer Evaluation}
For each surviving question with a non-empty answer, the second gate combines format checks with an evidence-appropriate reliability signal; representative accepted pairs are shown in Fig.~\ref{fig:example}:
\begin{equation}
A_i=\eta H_a(q_i,a_i)+(1-\eta)S_a(q_i,a_i),
\label{eq:answer_score}
\end{equation}
Here $H_a=(h_{\mathrm{type}}+h_{\mathrm{concise}}+h_{\mathrm{nonempty}})/3$ measures type validity, conciseness, and non-emptiness; $S_a$ averages correctness, relevance, completeness, and visual consistency; and $\eta$ is a fixed mixing coefficient. Refusal or uncertainty triggers hard rejection. Recognition and understanding use QA-pool evaluator judgments, whereas perception uses type-aware pseudo-label matching, with a mismatch setting $G_i=0$. Acceptance requires $A_i\geq\tau^a_{d_i}$ and no hard rejection. This evidence-matched gate lets deterministic and semantic supervision follow one training objective while retaining their distinct verification signals.

\paragraph{Threshold Adaptation}
As competence and score distributions change, fixed thresholds impose unsuitable selection pressure. CurEvo instead adapts question-type and dimension-level thresholds. Let $z$ index either case and $d(z)$ its associated dimension:
\begin{equation}
\begin{aligned}
p_{d(z)}^{(t)}&=\operatorname{clip}\left(p_0+\gamma c_{d(z)}^{(t)},p_{\min},p_{\max}\right),\\
\widetilde{\tau}_{z}^{(t+1)}&=\operatorname{Quantile}_{p_{d(z)}^{(t)}}\left(V_z^{(t)}\right),\\
\tau_{z}^{(t+1)}&=\operatorname{clip}\left(\beta\tau_z^{(t)}+(1-\beta)\widetilde{\tau}_{z}^{(t+1)},\tau_{\min},\tau_{\max}\right).
\end{aligned}
\label{eq:threshold_update}
\end{equation}
Here $p_0$ is the base quantile, $\gamma$ scales the competence feedback, $\beta$ controls temporal smoothing, and the $p$ and $\tau$ bounds limit abrupt changes. For question thresholds, $V_z^{(t)}$ includes every score of type $z$; for answer thresholds, it includes every non-empty answer score in dimension $d$, including later rejections. The competence signal $c_d^{(t)}$ is mean correctness from pseudo-label matching for perception and semantic evaluation otherwise. Higher competence selects a higher quantile, while lower competence retains broader supervision. Empty score sets preserve the previous threshold, with the exact configuration given in the supplementary material. Because rejected records remain in the score sets, the threshold state reflects the quality distribution encountered in the current round rather than only the accepted tail. The updated threshold governs the next admission decision, while answer-correctness feedback also returns to the allocation update, so evaluation provides both local selection and global curriculum feedback.

\textbf{Training and Inference} After each iteration, LoRA adapts the base model on the retained pool $\mathcal{P}_{t+1}$ using Eq.~\ref{eq:final_loss}, while the QA-pool evaluator remains frozen and provides the selection feedback used to update the next curriculum state. The resulting adapter is carried into the following iteration, completing the recurrent update; at inference, the evolved base model answers questions directly without either evaluator or online update. Backend-specific collators and effective-weight implementations are provided in the supplementary material.

\begin{table*}[!t]
\centering
\caption{Main results on four VideoQA benchmarks. Each backbone is evaluated before and after CurEvo with Qwen2.5-VL(7B) as the frozen QA-pool evaluator. Accuracy follows each dataset's protocol, while the fixed Qwen2.5-VL(7B) benchmark scoring evaluator computes the 1--5 semantic Score. Parentheses give gains over the vanilla model (percentage points for Accuracy and score points for Score). Best and second-best results are bold and underlined, respectively.}
\label{tab:benchmark}
\small
\resizebox{\linewidth}{!}{
\begin{tabular}{l cc cc cc cc}
\toprule
\multirow{2}{*}{\textbf{Model}} &
\multicolumn{2}{c}{\textbf{ActivityNet-QA}} &
\multicolumn{2}{c}{\textbf{MSRVTT-QA}} &
\multicolumn{2}{c}{\textbf{MSVD-QA}} &
\multicolumn{2}{c}{\textbf{NExT-QA}} \\
\cmidrule(lr){2-3} \cmidrule(lr){4-5} \cmidrule(lr){6-7} \cmidrule(lr){8-9}
& \textbf{Accuracy (\%)}$\uparrow$ & \textbf{Score}$\uparrow$
& \textbf{Accuracy (\%)}$\uparrow$ & \textbf{Score}$\uparrow$
& \textbf{Accuracy (\%)}$\uparrow$ & \textbf{Score}$\uparrow$
& \textbf{Accuracy (\%)}$\uparrow$ & \textbf{Score}$\uparrow$ \\
\midrule
Video-LLaVA(7B)
& 45.36 & 3.31
& 34.40 & 2.84
& 40.20 & 3.01
& 57.14 & 3.77 \\
Video-LLaVA(7B) + \textbf{CurEvo}
& 48.72 (+3.36) & 3.45 (+0.14)
& 37.10 (+2.70) & 3.02 (+0.18)
& 43.00 (+2.80) & 3.18 (+0.17)
& 61.02 (+3.88) & 3.96 (+0.19) \\

LLaVA-OneVision(7B)
& 58.47 & 2.53
& 38.20 & 2.88
& 43.90 & 3.03
& 66.07 & 3.93 \\
LLaVA-OneVision(7B) + \textbf{CurEvo}
& 62.39 (+3.92) & 2.88 (+0.35)
& 41.80 (+3.60) & 3.12 (+0.24)
& 47.70 (+3.80) & 3.28 (+0.25)
& 70.12 (+4.05) & 4.09 (+0.16) \\

VILA(7B)
& 60.28 & 3.21
& 39.60 & 2.96
& 45.30 & 3.15
& 75.02 & 3.41 \\
VILA(7B) + \textbf{CurEvo}
& 62.47 (+2.19) & 3.37 (+0.16)
& 42.10 (+2.50) & 3.14 (+0.18)
& 47.90 (+2.60) & 3.32 (+0.17)
& 77.36 (+2.34) & 3.59 (+0.18) \\

Video-LLaMA3(7B)
& 62.04 & 3.49
& 36.80 & 2.92
& 42.60 & 3.08
& 79.38 & 4.37 \\
Video-LLaMA3(7B) + \textbf{CurEvo}
& 64.11 (+2.07) & 3.61 (+0.12)
& 39.90 (+3.10) & 3.13 (+0.21)
& 45.80 (+3.20) & 3.29 (+0.21)
& 80.86 (+1.48) & 4.44 (+0.07) \\

InternVL2.5(8B)
& 64.11 & 3.84
& 42.60 & 3.18
& 48.50 & 3.34
& 77.18 & 4.35 \\
InternVL2.5(8B) + \textbf{CurEvo}
& 66.02 (+1.91) & \underline{3.90} (+0.06)
& 44.80 (+2.20) & 3.33 (+0.15)
& 50.90 (+2.40) & 3.48 (+0.14)
& 79.46 (+2.28) & 4.43 (+0.08) \\

Qwen2.5-VL(7B)
& 67.84 & 3.72
& 41.80 & 3.10
& 47.60 & 3.26
& 76.94 & 4.28 \\
Qwen2.5-VL(7B) + \textbf{CurEvo}
& 69.91 (+2.07) & 3.81 (+0.09)
& 44.20 (+2.40) & 3.28 (+0.18)
& 50.10 (+2.50) & 3.43 (+0.17)
& 79.61 (+2.67) & 4.39 (+0.11) \\

Qwen3-VL(8B)
& \underline{70.41} & 3.88
& \underline{46.20} & \underline{3.48}
& \underline{52.10} & \underline{3.56}
& \underline{82.17} & \underline{4.45} \\
Qwen3-VL(8B) + \textbf{CurEvo}
& \textbf{71.02} (+0.61) & \textbf{3.93} (+0.05)
& \textbf{47.05} (+0.85) & \textbf{3.56} (+0.08)
& \textbf{52.95} (+0.85) & \textbf{3.64} (+0.08)
& \textbf{82.94} (+0.77) & \textbf{4.50} (+0.05) \\
\bottomrule
\end{tabular}
}
\end{table*}

\begin{table*}[t]
\centering
\caption{Cross-model evaluation on ActivityNet-QA. Rows vary the base backbone and columns denote separate CurEvo runs with the indicated QA-pool evaluator; the benchmark scoring evaluator is fixed to Qwen2.5-VL(7B). The Qwen2.5-VL(7B) column uses the default evaluator in Table~\ref{tab:benchmark}. Best and second-best results are bold and underlined, respectively.}
\label{tab:cross_model}
\small
\begin{tabular}{lcccc}
\toprule
\multirow{2}{*}{\textbf{Model}}
& \multicolumn{1}{c}{\textbf{VILA(7B)}} 
& \multicolumn{1}{c}{\textbf{InternVL2.5(8B)}} 
& \multicolumn{1}{c}{\textbf{Qwen2.5-VL(7B)}} 
& \multicolumn{1}{c}{\textbf{Qwen3-VL(8B)}} \\
\cmidrule(lr){2-2} \cmidrule(lr){3-3} \cmidrule(lr){4-4} \cmidrule(lr){5-5}
& \textbf{Accuracy (\%)}$\uparrow$ 
& \textbf{Accuracy (\%)}$\uparrow$ 
& \textbf{Accuracy (\%)}$\uparrow$ 
& \textbf{Accuracy (\%)}$\uparrow$ \\
\midrule
Video-LLaVA(7B) + \textbf{CurEvo}
& 47.30
& 47.97
& 48.72
& 49.15 \\

LLaVA-OneVision(7B) + \textbf{CurEvo}
& 61.10
& 61.76
& 62.39
& 62.86 \\

VILA(7B) + \textbf{CurEvo}
& 61.66
& 62.12
& 62.47
& 62.74 \\

Video-LLaMA3(7B) + \textbf{CurEvo}
& 62.92
& 63.45
& 64.11
& 64.34 \\

InternVL2.5(8B) + \textbf{CurEvo}
& 64.88
& 65.36
& 66.02
& 66.31 \\

Qwen2.5-VL(7B) + \textbf{CurEvo}
& \underline{67.28}
& \underline{68.24}
& \underline{69.91}
& \underline{70.26} \\

Qwen3-VL(8B) + \textbf{CurEvo}
& \textbf{69.44}
& \textbf{70.10}
& \textbf{71.02}
& \textbf{71.33} \\
\bottomrule
\end{tabular}
\end{table*}

\section{Experiments}
\label{sec:experiments}
We evaluate CurEvo from three complementary views: transfer across backbone families, coverage of distinct benchmark capabilities, and the contribution of quality selection and curriculum feedback. We first establish a common experimental protocol, then connect the main results and controlled analyses to the supervision trajectory produced by the framework.

\subsection{Experimental Setup}
\paragraph{Implementation Details}
For the main results, a separate frozen Qwen2.5-VL(7B) instance is the QA-pool evaluator for scoring and filtering generated pairs; an independent Qwen2.5-VL(7B) benchmark scoring evaluator computes the final 1--5 semantic Score. We evaluate Qwen3-VL~\cite{bai2025qwen3}, Qwen2.5-VL~\cite{bai2025qwen2}, InternVL2.5~\cite{chen2024expanding}, Video-LLaMA3~\cite{zhang2025videollama}, LLaVA-OneVision~\cite{li2024llava}, VILA~\cite{lin2024vila}, and Video-LLaVA~\cite{lin2024video}; Video-STaR and STEP provide controlled ActivityNet-QA baselines. All models use LoRA~\cite{hu2022lora, wang2024tr} under six iterations (seed 42, eight frames, and playback speeds $\{0.75,1.0,1.25\}$). Training videos are interleaved after task-specific QA annotations are removed, with validation and test splits reserved; initial question and answer thresholds are 0.62. Each iteration uses six epochs, learning rate $2\times10^{-5}$, cosine decay~\cite{vaswani2017attention}, warmup ratio 0.1, and mixed precision~\cite{micikevicius2017mixed}; backend settings are in the supplement. The QA-pool evaluator is frozen within each run, while the benchmark scoring evaluator is called only after prediction; neither is used at inference.

\paragraph{Benchmarks}
We evaluate on ActivityNet-QA~\cite{yu2019activitynet}, NExT-QA~\cite{xiao2021next}, MSRVTT-QA~\cite{xu2017video}, and MSVD-QA~\cite{xu2017video}. MSRVTT-QA and MSVD-QA emphasize short-video grounding and fine-grained recognition, where answers depend on local visual evidence. ActivityNet-QA stresses longer activity-centered videos and multi-stage events, while NExT-QA focuses on temporal dependencies and causal relations that require evidence integration across time. The four benchmarks therefore expose complementary demands, from direct visual evidence and semantic recognition to long-range event integration and higher-level reasoning, matching the three supervision dimensions used by CurEvo.

\paragraph{Metrics}
We report dataset-specific accuracy under each adopted matching protocol: ActivityNet-QA uses exact matching with the implementation's semantic-equivalent handling, NExT-QA uses option accuracy, and MSRVTT-QA/MSVD-QA use normalized short-answer matching. We also report a 1--5 benchmark semantic score, assigning strict matches the maximum and evaluating non-strict matches with the fixed benchmark scoring evaluator. Semantic equivalence changes correctness only for ActivityNet-QA; the other datasets retain strict accuracy. The semantic score grades correctness, relevance, completeness, and visual consistency, and is separate from the 0--1 QA-pool scores.

\begin{table*}[t]
\centering
\caption{Question-dimension ablation on four benchmarks. We remove perception, recognition, and understanding individually and in pairs. Best and second-best results are shown in bold and underlined, respectively.}
\label{tab:dimension_ablation}
\scriptsize
\setlength{\tabcolsep}{2.5pt}
\begin{tabular}{l cc cc cc cc}
\toprule
\multirow{2}{*}{\textbf{Model Variant}} &
\multicolumn{2}{c}{\textbf{ActivityNet-QA}} &
\multicolumn{2}{c}{\textbf{MSRVTT-QA}} &
\multicolumn{2}{c}{\textbf{MSVD-QA}} &
\multicolumn{2}{c}{\textbf{NExT-QA}} \\
\cmidrule(lr){2-3} \cmidrule(lr){4-5} \cmidrule(lr){6-7} \cmidrule(lr){8-9}
& \textbf{Accuracy (\%)}$\uparrow$ & \textbf{Score}$\uparrow$
& \textbf{Accuracy (\%)}$\uparrow$ & \textbf{Score}$\uparrow$
& \textbf{Accuracy (\%)}$\uparrow$ & \textbf{Score}$\uparrow$
& \textbf{Accuracy (\%)}$\uparrow$ & \textbf{Score}$\uparrow$ \\
\midrule
Qwen2.5-VL(7B) + \textbf{CurEvo} (full)
& \textbf{69.91} & \textbf{3.81}
& \textbf{44.20} & \textbf{3.28}
& \textbf{50.10} & \textbf{3.43}
& \textbf{79.61} & \textbf{4.39} \\
\midrule
w/o Perception
& 67.88 & 3.74
& \underline{41.70} & \underline{3.16}
& \underline{47.30} & \underline{3.29}
& 76.92 & 4.28 \\
w/o Recognition
& \underline{68.35} & \underline{3.76}
& 40.90 & 3.14
& 46.80 & 3.26
& \underline{77.36} & \underline{4.30} \\
w/o Understanding
& 67.41 & 3.73
& 41.20 & 3.15
& 47.00 & 3.27
& 76.58 & 4.24 \\
\midrule
w/o Perc. + Rec.
& 65.72 & 3.66
& 38.60 & 3.02
& 44.40 & 3.13
& 74.83 & 4.18 \\
w/o Perc. + Und.
& 65.09 & 3.63
& 38.10 & 2.99
& 43.90 & 3.10
& 74.21 & 4.14 \\
w/o Rec. + Und.
& 65.36 & 3.64
& 38.40 & 3.00
& 44.10 & 3.11
& 74.46 & 4.15 \\
\bottomrule
\end{tabular}
\end{table*}

\begin{table}[t]
\centering
\caption{Component ablation on ActivityNet-QA. The table compares generated QA, QA-pool-evaluator-guided selection, iterative evolution, and the adaptive curriculum controls. Video-STaR and STEP are included as external self-evolution baselines.}
\label{tab:ablation}
\scriptsize
\setlength{\tabcolsep}{3.5pt}
\renewcommand{\arraystretch}{1.05}
\begin{tabular}{lcc}
\toprule
\textbf{Variant} & \textbf{Accuracy (\%)} & \textbf{Score} \\
\midrule
Vanilla base (Qwen2.5-VL(7B)) & 67.84 & 3.72 \\
\midrule
+ Gen. QA & 62.48 & 3.40 \\
+ QA Filter & 64.12 & 3.48 \\
+ Iter. Evo. & 65.96 & 3.61 \\
Video-STaR~\cite{zohar2024video} & 68.52 & 3.76 \\
STEP~\cite{qiu2025step} & 68.78 & 3.77 \\
\midrule
w/o Adaptive Ratio & 67.26 & 3.69 \\
w/o Adaptive Threshold & 68.41 & 3.75 \\
\textbf{CurEvo} & \textbf{69.91} & \textbf{3.81} \\
\bottomrule
\end{tabular}
\end{table}

\begin{figure}[t]
    \centering
    \includegraphics[width=\linewidth]{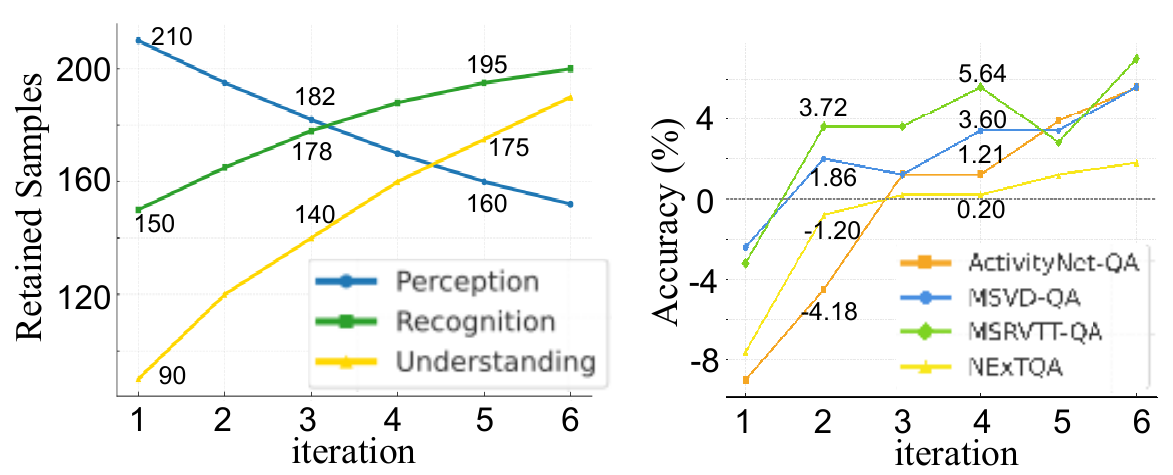}
    \caption{Training dynamics of CurEvo. Left: the retained supervision composition across self-evolution iterations. Right: accuracy change relative to the corresponding vanilla model as training proceeds.}
    \label{fig:iteration}
\end{figure}

\subsection{Main Results}
\label{sec:main_results}
Table~\ref{tab:benchmark} shows consistent gains across seven backbones and four benchmarks. For Qwen2.5-VL(7B), CurEvo raises ActivityNet-QA accuracy from 67.84\% to 69.91\% and NExT-QA accuracy from 76.94\% to 79.61\%; the corresponding semantic scores rise from 3.72 to 3.81 and from 4.28 to 4.39. Improvements also extend to short-video benchmarks and Qwen3-VL(8B).

\textbf{MSRVTT-QA and MSVD-QA} These benchmarks emphasize short-video grounding and recognition, where the model must align visible entities, actions, and attributes with concise answers. CurEvo improves both accuracy and semantic score across the compared backbones, supporting complementary perception and recognition supervision for local visual--language alignment. The parallel gains indicate improvements in both exact task performance and answer quality under the adopted evaluation protocols.

\textbf{ActivityNet-QA} ActivityNet-QA stresses longer activity-centered videos and multi-stage event understanding. Its gains show that the recurrent curriculum supports integration over a broader temporal span, connecting directly grounded cues with event-level interpretation. Since every vanilla row uses the same pretrained backbone as its CurEvo counterpart, the paired differences isolate the contribution of the self-evolution process rather than model capacity.

\textbf{NExT-QA} NExT-QA focuses on temporal dependencies, event ordering, and causal relations. Its consistent improvement provides evidence that CurEvo extends beyond perceptual grounding and recognition to higher-level video reasoning. The consistent direction across model families and benchmark types shows that quality-selective curriculum control complements, rather than replaces, the underlying Video-LLM.

\textbf{Cross-Model Generalization} Table~\ref{tab:cross_model} reports ActivityNet-QA results from separate CurEvo runs with different QA-pool evaluators; the benchmark scoring evaluator remains fixed to Qwen2.5-VL(7B). Backbone ordering stays stable across evaluator conditions, although absolute accuracy shifts with supervision selection. This indicates that CurEvo's supervision is not tied to one evaluator--base pairing: the QA-pool evaluator shapes self-evolution, while the adapted backbone retains the resulting capability under a common scoring protocol.

\subsection{Ablation and Analysis}
We compare controlled self-training variants, isolate question dimensions, and inspect iteration dynamics to link benchmark gains to selection, capability composition, and recurrent training. The component study follows the loop from candidate construction to retained training, the dimension study tests the capability composition that the loop maintains, and the iteration study examines how competence feedback changes supervision over time.

\paragraph{Component Ablation}
Table~\ref{tab:ablation} compares generated QA, QA-pool-evaluator filtering, iterative evolution, and the curriculum controls. Performance rises from self-generated to filtered, reused supervision; the complete configuration reaches 69.91\% and 3.81 on ActivityNet-QA. It exceeds external self-evolution baselines and variants without either control, supporting coordinated selection beyond generic self-training. This staged progression links cleaner supervision and repeated reuse to the final curriculum-controlled update.

\paragraph{Question-Dimension Ablation}
Table~\ref{tab:dimension_ablation} removes each question dimension individually and in pairs. Every removal lowers results, and paired removals lower them further. Recognition is especially useful for short videos, while perception and understanding provide grounding and temporal structure for activity and causal questions, supporting their combination in one curriculum.

The ablation pattern explains why the dimensions are generated and evaluated together. No single branch supplies the direct cues, semantic content, and cross-event relations needed across all four benchmarks. Their joint use allows the allocation state to respond to different competence gaps while the QA-pool evaluator preserves a shared acceptance path.

\paragraph{Iteration Analysis}
Figure~\ref{fig:iteration} summarizes retained supervision and accuracy. CurEvo recomputes allocation from the competence profile, allowing each dimension's share to rise or fall, while generation, deduplication, and quality gates preserve coverage. Accuracy rises overall with iteration-specific variation, reflecting feedback-driven adjustment. Because allocation follows current-round correctness rather than a fixed schedule, the left panel reflects evolving competence; cumulative weighting keeps accepted pairs available alongside newly targeted examples. Together, the panels show a responsive, cumulative curriculum.

\section{Conclusion}
\label{sec:conclusion}
CurEvo organizes self-generated VideoQA supervision as a competence-conditioned loop: multi-dimensional generation creates complementary candidates, type-adaptive evaluation retains evidence-supported pairs, and cumulative weighted training transfers them to the next state. Across seven backbones and four benchmarks, CurEvo improves accuracy and semantic score; for Qwen2.5-VL(7B), gains reach 2.07 and 2.67 percentage points on ActivityNet-QA and NExT-QA. These results support curriculum-guided quality selection as a scalable route to self-evolution without task-specific human QA annotations. By coupling evidence-matched selection with competence-conditioned allocation, CurEvo turns unlabeled videos into reusable supervision aligned with changing model needs.

\bibliographystyle{IEEEtran}
\bibliography{main}

\clearpage
\markboth{CurEvo: Supplementary Material}{CurEvo: Supplementary Material}
\section*{Supplementary Material}
\appendices
\fontencoding{T1}\selectfont
\setlength{\textfloatsep}{8pt plus 2pt minus 2pt}
\setlength{\floatsep}{6pt plus 2pt minus 2pt}
\setlength{\intextsep}{7pt plus 2pt minus 2pt}
\lstset{basicstyle=\footnotesize\ttfamily,columns=fullflexible,
  breaklines=true,breakatwhitespace=true,showstringspaces=false,
  keepspaces=true,frame=single,framerule=0.3pt,framesep=4pt,
  aboveskip=7pt,belowskip=7pt,xleftmargin=4pt,xrightmargin=4pt,
  upquote=true}

\raggedbottom
This supplementary material details the implementation of CurEvo for \emph{Curriculum-Guided Self-Evolution for Video Understanding}. It specifies question construction, video allocation, quality assessment, adaptive thresholds, cumulative training, and benchmark scoring. Complete prompt templates and worked calculations make the protocol directly reproducible. The settings below follow the paper configuration and its formal Qwen2.5-VL(7B) configuration. Table and equation numbering continues from the main paper.

The learner generates recognition and understanding questions and answers; controlled video transformations supply perception questions and deterministic targets. A frozen QA-pool evaluator assesses question quality and answer reliability. A separate benchmark scoring evaluator assesses benchmark predictions after inference. For the main configuration, both evaluator roles use frozen Qwen2.5-VL(7B) instances. Their inputs and outputs are specified separately below.

\section{Question Type Mapping}
\label{sec:supp_mapping}
CurEvo assigns each candidate to one of twelve question types within three capability dimensions. This assignment controls the generation guidance, answer parser, correctness signal, and threshold lookup. Table~\ref{tab:supp_type_mapping} gives the complete mapping. All four types in the assigned dimension are requested for each video, with one question per type.

\begin{table*}[t]
\caption{Complete question-type mapping. Perception questions use controlled sampled-frame relations; recognition and understanding questions use evidence-conditioned generation.}
\label{tab:supp_type_mapping}
\centering
\footnotesize
\setlength{\tabcolsep}{5pt}
\renewcommand{\arraystretch}{1.13}
\begin{tabularx}{\textwidth}{p{0.12\textwidth}p{0.12\textwidth}YY}
\toprule
\textbf{Dimension} & \textbf{Type} & \textbf{Question content and visual evidence} & \textbf{Answer representation and verification}\\
\midrule
Perception & Counting & Number of sampled frames presented to the model. & Integer; matched to the sampled-index count.\\
 & Localization & Ordinal position of a sampled frame marked with a red border. & Ordinal or integer; matched to the marked position.\\
 & Disorder & Whether the presented sequence preserves its original chronological order. & Yes/no; matched to the recorded permutation.\\
 & Consistency & Whether the sequence is continuous without an exact duplicated neighboring frame. & Yes/no; matched to the duplication operation.\\
\midrule
Recognition & Object & A clearly visible object, including an object held by a person. & Short factual answer; semantic assessment.\\
 & Action & The main visible activity or action. & Short factual answer; semantic assessment.\\
 & Attribute & Visible color, shape, size, or appearance. & Short factual answer; semantic assessment.\\
 & Scene & The setting or location of the visible event. & Short factual answer; semantic assessment.\\
\midrule
Understanding & Temporal & Event order, before/after relations, or temporal progression. & Evidence-supported answer; semantic assessment.\\
 & Spatial & A visible spatial relation between entities. & Evidence-supported answer; semantic assessment.\\
 & Causal & A cause inferable from the shown evidence. & Evidence-supported answer; semantic assessment.\\
 & Intention & A person's goal or intention supported by the video. & Evidence-supported answer; semantic assessment.\\
\bottomrule
\end{tabularx}
\end{table*}

The perception types refer to the sequence actually shown to the learner. For example, counting asks for the number of sampled frames, and localization asks for the marked frame's ordinal position. Their labels are computed from the view specification. Recognition questions identify visible content, while understanding questions query relations among entities, events, causes, and goals. Temporal and spatial types are assigned by the relation being queried; a directly visible relation is categorized by that relation. This rule keeps type assignment consistent across simple and more compositional examples.

For generated NExT-QA-style questions, recognition and understanding retain the same semantic types but include a short choice list. The visible evidence supports one option, and the remaining options serve as distractors. The option parser and choice-specific checks supplement the common semantic assessment.

\section{Generation and Normalization}
\label{sec:supp_generation}
\subsection{Controlled Perception Questions}
Each perception record contains a question, a view specification, and a deterministic pseudo label. The view records the sampled indices, temporal speed, and any marking, permutation, or duplication. These fields connect the label to the same sequence supplied to the learner and evaluator.

\subsubsection{Counting}
The fixed question is ``How many sampled frames are shown in this sequence?''\ The label is the number of sampled indices. With the default eight-frame input, the target is eight; a source with fewer than eight frames uses all available indices and the corresponding count.

\subsubsection{Localization}
The question is ``What is the ordinal position of the sampled frame with a red border?''\ For a sequence of length $K$, the zero-based marked index is $\min(1,K-1)$. Thus, the second sampled frame is marked when $K\geq2$, and the only frame is marked when $K=1$. The target is its one-based position. The border is red, with RGB value $(255,32,32)$, and its width is $\max(3,\lfloor224/40\rfloor)$ pixels under the common image reader.

\subsubsection{Disorder}
The question is ``Are these sampled frames presented in their original chronological order?''\ The generator alternates unchanged and transformed cases for this type. For $K\geq4$, the transformed case swaps positions 1 and $K-2$ in zero-based indexing. For two or three frames, it reverses the sequence. A nonidentity permutation gives the target ``no''; an identity permutation gives ``yes.''\ The label is therefore determined by the actual index order.

\subsubsection{Consistency}
The question is ``Is this sampled sequence continuous without an exact duplicated neighboring frame?''\ This type also alternates transformed and unchanged cases. When $K\geq2$, the transformed case replaces frame $\min(2,K-1)$ with a copy of its predecessor. The label is ``no''\ when this operation is applied and ``yes''\ for the unchanged view. The duplication concerns the sampled input sequence and is recorded explicitly in the view.

Together, these constructions provide observable supervision for frame quantity, positional grounding, temporal order, and neighboring-frame consistency. The learner still answers each question from the presented images. Its answer is subsequently checked against the recorded pseudo label.

\subsection{Recognition and Understanding Questions}
For recognition and understanding, the current learner inspects the sampled frames and receives the requested dimension, type list, and type-specific guidance. It generates questions before answering them. The prompt requests concise English questions grounded in visible evidence and returns a JSON object containing a list of typed questions. Section~\ref{sec:supp_prompts} reproduces the templates and all eight semantic type instructions.

The paper configuration requests one question for each of the four types in a dimension. Question generation uses temperature 0.2, with a maximum of 256 new tokens for ordinary questions and 512 for questions with choices. The larger choice budget accommodates the option arrays. Each question that passes question assessment is then answered by the current learner with a maximum of 192 new tokens. The answer prompt requests a short factual phrase, exactly ``yes''\ or ``no''\ for binary forms, or ``LETTER. option text''\ for a choice-based item.

\subsection{Structured Output and Question Checks}
The JSON interface accepts a JSON object, a fenced JSON response, or a top-level list wrapped as a question list. Trailing commas outside quoted strings are removed before parsing. The interface permits two repair retries, for at most three model calls. A repair instruction requests valid JSON without Markdown while preserving the original task.

The generator then applies the following checks:
\begin{enumerate}
\item Retain object records whose type is known and belongs to the requested type list.
\item Collapse repeated whitespace and discard empty questions.
\item Append a final question mark when one is absent, and discard questions exceeding 40 whitespace-delimited words.
\item Enforce the requested per-type count, which is one in the paper configuration, and remove near-duplicate questions within the same video and type.
\end{enumerate}
For deduplication, question text is lowercased, punctuation is removed in favor of alphanumeric text and spaces, and whitespace is collapsed. A candidate is redundant when its \texttt{SequenceMatcher} similarity to a retained question reaches 0.88. This comparison is local to a video and type, preserving distinct supervision across types and videos.

\subsection{Choice Construction and Answer Canonicalization}
Generated choice-based items use three to five options. Each option must be nonempty, contain at most 16 words, and be distinct after normalization. The prompt requires one visually supported option and plausible distractors, without revealing the answer. The valid choices are shuffled with the experiment seed before both the learner and evaluator receive them. Consequently, their option letters refer to the same shuffled list.

The learner's answer must map to a legal option. The parser recognizes a standalone letter, a letter followed by punctuation and option text, forms such as ``option A''\ or ``answer: A,''\ and an exact normalized option string. The evaluator also returns an independently inferred option in its assessment record. If that answer cannot be parsed as a legal choice, the record receives a hard rejection reason.

For training, option answers are converted to their canonical letters. Counting and localization use numeric targets, while disorder and consistency use ``yes''\ or ``no.''\ Ordinary answers are lowercased and normalized by removing punctuation, the articles ``a,''\ ``an,''\ and ``the,''\ and redundant spaces. Integer parsing handles digit forms and English numbers from zero to twelve; ordinal parsing handles the first through eighth positions and numeric forms. The original target is retained alongside the canonical target, preserving the association between the generated wording and the tokenized supervision.

\section{Sampling Protocol}
\label{sec:supp_sampling}
\subsection{Video Sources and Assignment Budget}
The self-evolution pool draws videos from the training splits of ActivityNet, MSRVTT, MSVD, and NExT-QA. Task-specific QA annotations are excluded from generation and model updates. Validation and test QA records are reserved for evaluation.

Each iteration makes 90 video assignments. Initial dimension ratios are $(0.50,0.30,0.20)$ for perception, recognition, and understanding, respectively, giving 45, 27, and 18 assignments in the first iteration. For subsequent iterations, the sampler multiplies each ratio by 90, takes the integer parts, and allocates the remaining assignments to the largest fractional remainders. The resulting dimension counts sum to 90.

Within each dataset, the video list is shuffled with seed 42 and consumed sequentially. Dataset keys are sorted, and the sampler draws from them in round-robin order while filling the dimension counts. A dataset list is reshuffled when exhausted. Finally, the complete assignment list is shuffled before generation. This procedure interleaves the four sources while allowing the curriculum to determine dimension coverage.

Four types per assignment and one question per type yield a nominal budget of $90\times4=360$ candidate requests per iteration. Across six iterations, this is at most 2,160 requested candidates before parsing, deduplication, and quality filtering. These values describe the configured request budget; realized candidate and accepted counts are saved separately.

\subsection{Frame Selection and Temporal Variation}
The common video reader uses eight sampled frames resized to $224\times224$ pixels with bicubic interpolation. If a source has $F\leq8$ frames, all frame indices are used. Otherwise, let $K=8$ and let the sampled temporal-speed factor be $v\in\{0.75,1.0,1.25\}$. The temporal interval is constructed as
\begin{equation}
\begin{aligned}
b&=\clip(0.8v,0.25,1),\\
s&=\max\{K-1,\lfloor(F-1)b\rfloor\},\\
a&=\max\{0,\lfloor(F-1-s)/2\rfloor\},\\
e&=\min\{F-1,a+s\}.
\end{aligned}
\label{eq:supp_frame_span}
\end{equation}
The $K$ indices are obtained by integer-valued linear spacing from $a$ to $e$, using \texttt{numpy.linspace} with integer output. Thus, the speed setting changes the temporal span covered by a fixed number of images. Frames are converted to RGB before resizing. Perception marking, permutation, or duplication is applied to this sampled view, and the same view specification is reused during answering, scoring, and training.

\subsection{Competence-Conditioned Allocation}
For each dimension $d$, current-round competence $c_d^{(t)}$ is the mean answer-correctness field over records with nonempty learner answers. Perception contributes deterministic match values; recognition and understanding contribute evaluator correctness. An empty dimension receives competence zero. The dimension ratios follow
\begin{equation}
\begin{aligned}
u_d&=r_d^{(t)}[1+\lambda_d(1-c_d^{(t)})],\\
\bar r_d&=u_d/\sum_{d'}u_{d'},\\
r_d^{(t+1)}&=
\frac{\max(\rho,\bar r_d)}
{\sum_{d'}\max(\rho,\bar r_{d'})}.
\end{aligned}
\label{eq:supp_ratio}
\end{equation}
The sensitivities are $\lambda_d=0.50,0.75,1.00$ in perception--recognition--understanding order, and $\rho=0.10$. The floor is applied to the intermediate normalized ratios before the final normalization.

The multiplicative update increases a dimension's unnormalized allocation according to its measured competence gap and sensitivity. Final shares depend on all three dimensions, so a share may rise or fall between iterations. These ratios determine video assignments before generation. The accepted-pair distribution additionally depends on question validity, deduplication, format checks, and both quality gates. Saving assignment counts and accepted counts separately makes these two stages explicit.

\section{Type-Adaptive Evaluation Details}
\label{sec:supp_eval}
\subsection{Question Score}
The question gate combines a structural heuristic with visual-semantic assessment. For a question $q$, let $L$ denote its whitespace-delimited word count after whitespace normalization. The heuristic is
\begin{equation}
H_q(q)=
\frac{h_{\mathrm{len}}+h_{\mathrm{form}}+
h_{\mathrm{shape}}+h_{\mathrm{clear}}}{4}.
\label{eq:supp_hq}
\end{equation}
Its components are specified as follows.
\begin{enumerate}
\item \emph{Length:} $h_{\mathrm{len}}=1$ for $5\leq L\leq30$; otherwise, $h_{\mathrm{len}}=\max(0,1-|L-16|/24)$.
\item \emph{Question form:} $h_{\mathrm{form}}=1$ if the text contains exactly one question mark and ends with it; otherwise, it is 0.4.
\item \emph{Opening form:} $h_{\mathrm{shape}}=1$ for a leading word in \{what, who, where, when, why, how, which, is, are, does, do, did, has, have, can\}; otherwise, it is 0.5.
\item \emph{Clarity:} Substring matching sets $h_{\mathrm{clear}}=0$ when the lowercased question contains ``ground truth,''\ ``annotation,''\ ``dataset,''\ ``language model,''\ or ``cannot see''; otherwise, it is one.
\end{enumerate}

The frozen question evaluator receives the sampled frames, dimension, type, and candidate question. It returns three scores: semantic alignment, visual grounding, and answerability, together with a brief reason. Semantic alignment checks the requested cognitive type, visual grounding checks support in the frames, and answerability checks whether the shown evidence suffices for a factual answer. The evaluator is instructed to assess the question without answering it.

Each returned field is converted to a finite value in $[0,1]$: missing, nonnumeric, or nonfinite values become zero, and finite values are clipped. Let $S_q$ be the mean of the three sanitized fields. The combined score is
\begin{equation}
Q=0.35H_q+0.65S_q.
\label{eq:supp_qscore}
\end{equation}
The paper configuration enables a heuristic fallback when all three sanitized semantic fields equal zero. In that case, all three fields are replaced by $H_q$, yielding $S_q=H_q$ and $Q=H_q$; the record stores a fallback flag. The component values and evaluator reason are retained with the resulting score.

The question is selected when $Q\geq\tau^q_k$, where $k$ is its question type. Answer generation is performed only for selected questions. This ordering makes question validity an explicit prerequisite for generating a training answer.

\subsection{Answer Score}
The learner answers the selected question using the same sampled view. Its answer heuristic is
\begin{equation}
H_a(q,a)=
\frac{h_{\mathrm{type}}+h_{\mathrm{concise}}+
h_{\mathrm{nonempty}}}{3}.
\label{eq:supp_ha}
\end{equation}
Here $h_{\mathrm{type}}$ is one when the answer passes its type-specific parser and format gate, and zero otherwise. The nonempty indicator uses the stripped answer text. For normalized answer length $L_a$, the conciseness component is one for $1\leq L_a\leq12$ and otherwise follows
\begin{equation}
h_{\mathrm{concise}}=
\max\left(0,1-\frac{\max(0,L_a-12)}{28}\right).
\label{eq:supp_concise}
\end{equation}
Empty answers are rejected separately.

For recognition and understanding, the frozen answer evaluator receives the frames, question, optional choices, and learner answer. Its prompt first requests an independently inferred answer, followed by correctness, relevance, completeness, and visual consistency scores. These fields measure the answer's agreement with the evidence, its relevance to the question, its coverage of the requested information, and its consistency with the shown video. After the same finite-value sanitization, their mean defines $S_a$. The answer score is
\begin{equation}
A=0.35H_a+0.65S_a.
\label{eq:supp_ascore}
\end{equation}
The independently inferred answer supports assessment and is stored as metadata. In the paper configuration, accepted recognition and understanding targets remain the learner's own answers, as specified in Section~\ref{sec:supp_pool}-B.

For perception, let $m\in\{0,1\}$ be the type-aware match between the learner answer and the deterministic pseudo label, and let $v\in\{0,1\}$ denote format validity. The four semantic fields are $(m,v,m,m)$ in correctness--relevance--completeness--visual-consistency order, giving
\begin{equation}
S_a=(3m+v)/4.
\label{eq:supp_perception_score}
\end{equation}
A pseudo-label mismatch triggers hard rejection. Its score remains in the record for competence estimation and threshold statistics, while the acceptance decision enforces deterministic correctness.

\subsection{Sequential Gates and Recorded Decisions}
A pair enters the accepted set only when it passes the question threshold, produces a nonempty learner answer, passes the answer-format checks, and meets the dimension-specific answer threshold. Format checks reject unparseable integer, ordinal, binary, or choice answers as applicable. Ordinary open-ended answers must contain between one and 40 normalized words. Refusal markers include ``cannot determine,''\ ``can't determine,''\ ``not enough information,''\ and ``as an AI.''\ Choice-based records additionally require a legal evaluator option, and perception records require a pseudo-label match.

For every generated record, the pipeline retains the video identity, dimension, type, view specification, question, choices, learner answer, pseudo label where applicable, component scores, aggregate scores, acceptance flag, and rejection reason. Accepted records also store the original and canonical targets. These fields distinguish question rejection, answer-format rejection, incorrect perception answers, and answer-threshold rejection.

\subsection{Adaptive Thresholds}
All twelve question-type thresholds and all three dimension-level answer thresholds start at 0.62. Let $V_z^{(t)}$ be the current score set for threshold $z$, and let $d(z)$ be its dimension. The update is
\begin{equation}
\begin{aligned}
p_z^{(t)}&=\clip(0.45+0.30c_{d(z)}^{(t)},0.05,0.95),\\
\widehat{\tau}_z^{(t)}&=
\operatorname{Quantile}_{p_z^{(t)}}(V_z^{(t)}),\\
\tau_z^{(t+1)}&=
\clip(0.70\tau_z^{(t)}+
0.30\widehat{\tau}_z^{(t)},0.50,0.90).
\end{aligned}
\label{eq:supp_threshold}
\end{equation}
Quantiles use NumPy's default linear interpolation. For sorted scores $x_0,\ldots,x_{n-1}$, let $h=(n-1)p$, $j=\lfloor h\rfloor$, and $\delta=h-j$. The quantile is $(1-\delta)x_j+\delta x_{\min(j+1,n-1)}$.

A question threshold uses every recorded question score of its type, including question-gate rejections. An answer threshold uses every recorded answer score in its dimension with a nonempty learner answer, including subsequent format or threshold rejections. The competence signal uses the correctness field over these nonempty learner answers. An empty score set leaves its previous threshold unchanged.

The gate at iteration $t$ uses the thresholds present at that iteration's start. Statistics from that round determine the thresholds and ratios for the following round. The quantile sets the selection pressure according to both the encountered score distribution and measured competence, while exponential smoothing carries forward the preceding state. The saved iteration summary includes the ratios before and after updating, competence values, thresholds, and retained counts by dimension.

\section{Pool, Targets, and Optimization}
\label{sec:supp_pool}
\subsection{Training Pool}
Accepted records are appended to a cumulative pool, so reliable pairs from earlier iterations remain available for subsequent model updates. The configured capacity is 4,000 records. Capacity control ranks stored training priorities if this limit is exceeded, after which weights are refreshed over the retained pool. Under the six-round request budget of at most 2,160 candidates, the capacity accommodates every accepted record.

For each retained record, the informativeness proxy uses the canonical target when available and the learner answer otherwise. Lowercasing and the regular expression \texttt{[a-z0-9]+} produce the token sequence $\mathcal{T}_i$ of length $n$. For $n>0$,
\begin{equation}
\begin{aligned}
U_{\mathrm{len}}&=\min(1,n/12),\\
U_{\mathrm{div}}&=
|\operatorname{Unique}(\mathcal{T}_i)|/n.
\end{aligned}
\label{eq:supp_info}
\end{equation}
Let $C_i$ be the token sequence after removing the stop words \{a, an, the, is, are, to, of, in, on, it, and\}. Repeated content tokens remain in this sequence. Specificity and the combined score are
\begin{equation}
\begin{aligned}
U_{\mathrm{spec}}&=
\frac{\sum_{w\in C_i}\mathbf{1}[|w|\geq5]}
{\max(1,|C_i|)},\\
U_i&=\clip(0.35U_{\mathrm{len}}+
0.35U_{\mathrm{div}}\\
&\hspace{3.8em}+0.30U_{\mathrm{spec}},0.10,1).
\end{aligned}
\label{eq:supp_info_spec}
\end{equation}
An empty token sequence receives zero. The nonempty lower bound preserves a positive informativeness contribution for concise numeric, binary, and option-letter targets.

For current dimension ratio $r_{d_i}^{(t)}$, the raw priority and normalized weight are
\begin{equation}
\widetilde w_i=r_{d_i}^{(t)}\max(0,A_i)U_i,\qquad
w_i=\frac{\widetilde w_i}{\sum_{j\in\mathcal P}\widetilde w_j}.
\label{eq:supp_weight}
\end{equation}
If the denominator is nonpositive for a nonempty pool, uniform weights are assigned. Recomputing these weights over the cumulative pool lets the current curriculum affect historical and newly accepted supervision in the same update.

\subsection{Target Construction}
Perception targets are derived from the controlled view. Recognition and understanding targets are the current learner's generated answers after successful assessment. The evaluator supplies quality fields and the information used for admission; its independently inferred answer is retained in the assessment metadata. The default paper configuration uses learner-generated targets for these two dimensions.

Canonicalization makes the output interface consistent across training and evaluation: counting uses a digit, localization uses a numeric position, binary questions use ``yes''\ or ``no,''\ and choice questions use a legal option letter. Ordinary textual targets use the normalized factual answer. The original target is preserved as metadata before canonicalization.

Each training example combines the sampled video with the question and any option list using the backbone's native multimodal chat interface. Prompt tokens and padding positions are assigned the loss-ignore label $-100$. Only target-answer tokens contribute to the autoregressive objective. This constructs supervision from accepted factual answers while preserving the corresponding visual evidence.

\subsection{Backbone-Aware Training}
The Qwen configuration uses LoRA rank 16, scaling factor 32, and dropout 0.05. Adaptation targets the attention projections \texttt{q\_proj}, \texttt{k\_proj}, \texttt{v\_proj}, and \texttt{o\_proj}, together with \texttt{gate\_proj}, \texttt{up\_proj}, and \texttt{down\_proj} in the feed-forward blocks. Each iteration trains for six epochs with AdamW, learning rate $2\times10^{-5}$, weight decay 0.01, cosine scheduling, warmup ratio 0.10, gradient clipping at 1.0, mixed precision, and gradient checkpointing. The default Qwen microbatch size is two, gradient accumulation is four, and maximum sequence length is 1536. Training starts once the cumulative pool contains at least 200 accepted records.

The previous iteration's adapter initializes the next model update. The QA-pool evaluator remains frozen throughout the run. Iteration-specific training shuffling uses the project seed plus the zero-based iteration index. The resulting adapter is used for question generation and answering in the following round.

\begin{table}[t]
\caption{Backbone-aware input and optimization interfaces. The common quality gates and cumulative pool are shared across these interfaces.}
\label{tab:supp_backend}
\centering
\footnotesize
\setlength{\tabcolsep}{3pt}
\renewcommand{\arraystretch}{1.12}
\begin{tabularx}{\columnwidth}{p{0.25\columnwidth}Y}
\toprule
\textbf{Backend} & \textbf{Training interface}\\
\midrule
Qwen2.5-VL & Microbatch 2; accumulation 4; maximum length 1536; weighted per-example token loss.\\
Video-LLaMA3 & Microbatch 1; native collator and model loss; sample weight applied to the model-returned loss.\\
InternVL & Microbatch 1; native tokenizer and collator; 448-pixel visual input adapter; sample-weighted native loss.\\
\bottomrule
\end{tabularx}
\end{table}

For Qwen, let $x_i$ denote the visual input and question prompt. The token losses for sample $i$ are first averaged over its valid target positions $M_i$:
\begin{equation}
\ell_i=-\frac{1}{|M_i|}
\sum_{u\in M_i}\log p_\Theta(y_{i,u}\mid y_{i,<u},x_i).
\label{eq:supp_token_loss}
\end{equation}
If the retained pool has $N$ examples, the effective sample weight is $e_i=Nw_i'$, where $w_i'$ includes the choice-label balancing below. The microbatch loss for batch $\mathcal B$ is
\begin{equation}
\mathcal L_{\mathcal B}=
\frac{1}{|\mathcal B|}\sum_{i\in\mathcal B}e_i\ell_i.
\label{eq:supp_batch_loss}
\end{equation}
Weights are applied after token averaging, so target length does not determine an example's total contribution. Multiplication by $N$ gives effective weights with mean one over the pool. The implementation preserves these weights inside a microbatch and divides each microbatch loss by the gradient-accumulation factor. Video-LLaMA3 and InternVL use batch size one with their native model-returned loss multiplied by the same effective sample weight.

\subsection{Choice-Label Weight Balancing}
The paper configuration balances the total training mass of represented option letters. Let $\mathcal P_{\mathrm{mc}}$ contain choice-based records, $W_{\mathrm{mc}}=\sum_{i\in\mathcal P_{\mathrm{mc}}}w_i$, and $W_l$ be the weight mass of option letter $l$. If $K_{\mathrm{lab}}\geq2$ letters are represented, the adjusted weights are
\begin{equation}
w_i'=w_i\frac{W_{\mathrm{mc}}}{K_{\mathrm{lab}}W_{l_i}},
\qquad i\in\mathcal P_{\mathrm{mc}}.
\label{eq:supp_choice_balance}
\end{equation}
All other record weights remain unchanged. Each represented letter then carries mass $W_{\mathrm{mc}}/K_{\mathrm{lab}}$, and the total mass assigned to choice-based records is preserved. Combined with seeded choice shuffling, this prevents a frequently generated option position from dominating the training objective.

\section{Reproducibility Protocol}
\label{sec:supp_protocol}
\subsection{End-to-End Iteration Order}
A complete round operates on the learner adapter, curriculum state, and cumulative pool from the preceding round:
\begin{enumerate}
\item Allocate 90 video assignments using the current dimension ratios and the seeded interleaved sampler.
\item Construct the sampled views and request one candidate for each of the four types in the assigned dimension.
\item Normalize and deduplicate candidates, assess question quality, and retain questions meeting their type thresholds.
\item Generate learner answers for surviving questions, assess their reliability, and apply answer-format, pseudo-label, and answer-threshold gates.
\item Canonicalize accepted targets, append accepted records to the cumulative pool, and refresh training weights.
\item Apply choice-label balancing and train the learner adapter on the weighted pool.
\item Use current-round correctness and score distributions to update ratios and thresholds for the next round, and save the state and adapter.
\end{enumerate}
Six rounds use project seed 42. The common protocol is summarized in Table~\ref{tab:supp_protocol}.

\begin{table*}[t]
\caption{Paper protocol and implementation settings. Token limits are maximum generated lengths; the candidate budget precedes all filtering.}
\label{tab:supp_protocol}
\centering
\footnotesize
\setlength{\tabcolsep}{5pt}
\renewcommand{\arraystretch}{1.10}
\begin{tabularx}{\textwidth}{p{0.26\textwidth}Y}
\toprule
\textbf{Element} & \textbf{Setting}\\
\midrule
Video sources & Training-split ActivityNet, MSRVTT, MSVD, NExT-QA; task-specific QA annotations excluded.\\
Evolution schedule & Six iterations; project seed 42; cumulative accepted-pair pool.\\
Assignment budget & 90 videos per iteration; four types per video; one question per type.\\
Initial curriculum & Perception 0.50; recognition 0.30; understanding 0.20.\\
Common visual reader & Eight frames; $224\times224$ bicubic resize; speeds $\{0.75,1.0,1.25\}$.\\
Question generation & Temperature 0.2; 256 tokens for ordinary questions; 512 for choice-based questions.\\
Question normalization & Maximum 40 words; within-video, within-type similarity threshold 0.88.\\
Choice construction & Three to five options; maximum 16 words per option; seeded option shuffle.\\
Learner answering & Maximum 192 tokens; factual-phrase, binary, or option-specific output form.\\
QA-pool evaluator & Frozen Qwen2.5-VL(7B); temperature 0.0; maximum 256 tokens per assessment.\\
Score mixing & Question heuristic weight $\alpha=0.35$; answer heuristic weight $\eta=0.35$.\\
Question fallback & All three sanitized semantic fields zero: substitute $H_q$ and record a flag.\\
Initial thresholds & All question types and answer dimensions start at 0.62.\\
Threshold update & Base quantile 0.45; competence scale 0.30; EMA 0.70; probability bounds $[0.05,0.95]$; threshold bounds $[0.50,0.90]$.\\
Ratio update & Sensitivities $(0.50,0.75,1.00)$; intermediate ratio floor 0.10.\\
Informativeness & Length/diversity/specificity weights $(0.35,0.35,0.30)$; nonempty lower bound 0.10.\\
Training pool & Minimum 200 records for training; capacity 4,000; reweighted cumulative supervision.\\
LoRA & Rank 16; scaling 32; dropout 0.05; attention and feed-forward projection modules.\\
Optimization & Six epochs per iteration; AdamW; learning rate $2\times10^{-5}$; weight decay 0.01; cosine decay; warmup 0.10; gradient clip 1.0.\\
Qwen batching & Microbatch 2; gradient accumulation 4; maximum sequence length 1536.\\
Memory settings & Mixed precision and gradient checkpointing; backend-native visual and loss interfaces.\\
Choice-label balancing & Equal total mass across represented option letters, preserving the choice subset's total weight.\\
Benchmark decoding & Greedy learner decoding at temperature 0.0; maximum 192 new tokens.\\
Benchmark semantic scoring & Fixed Qwen2.5-VL(7B); temperature 0.0; maximum 256 tokens; numeric 1--5 rubric.\\
\bottomrule
\end{tabularx}
\end{table*}

\subsection{Benchmark Accuracy and Semantic Score}
Benchmark predictions are generated by the evolved learner directly. The benchmark scoring evaluator is called after prediction and does not participate in learner inference or curriculum updates. The learner uses greedy decoding at temperature 0.0 for the formal benchmark protocol.

Accuracy follows the adopted dataset-specific matching procedure. MSRVTT-QA and MSVD-QA use normalized short-answer matching; NExT-QA uses strict option matching. Binary answers are compared by their parsed yes/no values, and numeric references use integer matching. ActivityNet-QA counts a prediction as correct when it strictly matches the normalized reference or receives a positive semantic-equivalence judgment. These matching rules are applied consistently to the vanilla and evolved model.

The semantic Score uses a 1--5 rubric. A strict match receives score five. For non-strict open-ended predictions, the fixed benchmark evaluator receives the video frames, question, reference answer, and prediction. It assigns five for a fully correct equivalent answer, four for an essentially correct answer with a minor omission, three for a partially correct but incomplete answer, two for a weakly related but substantially incorrect answer, and one for an incorrect or unrelated answer. The prompt requests \texttt{equivalent=true} for scores four and five and false for scores one through three. Binary references use deterministic comparison, with score five for a match and one otherwise.

The scoring interface defines a deterministic fallback for missing numeric or equivalence fields. For a nonbinary answer, let $R$ and $P$ be the normalized reference and prediction token sets, and let
\begin{equation}
o=\frac{|R\cap P|}{\max(1,|R|)}.
\label{eq:supp_overlap}
\end{equation}
The fallback score is $1+4o$, and fallback equivalence is $o\geq0.8$; an empty reference gives score one and false equivalence. A returned numeric score is clipped to $[1,5]$, and a returned Boolean equivalence field is used directly. An absent or wrongly typed field uses its corresponding fallback. Stored scoring-source information distinguishes strict or binary decisions, evaluator JSON, and deterministic token-overlap scoring. The full semantic-scoring prompt appears in Section~\ref{sec:supp_prompts}.

QA-pool scores and benchmark scores serve separate roles. Question and answer scores in $[0,1]$ control admission, training weights, and competence feedback. Benchmark scores in $[1,5]$ assess predictions against benchmark references after inference. Benchmark references remain within the evaluation procedure.

\subsection{Saved State and Reconstruction}
Each iteration saves generated records, accepted records, rejection reasons, the cumulative pool, curriculum statistics, and the learner adapter. The view specification reconstructs frame indices and controlled transformations. Record fields reconstruct score aggregation, threshold comparisons, answer canonicalization, and training weights. The curriculum snapshot preserves the ratios, competence values, question-type thresholds, and answer-dimension thresholds used to start the next round.

Generation and evaluation are recorded separately so that question survivors can be distinguished from accepted training pairs. Benchmark records likewise retain predictions, strict-match flags, semantic judgments, and scoring sources. These outputs connect each final measurement to its prediction and connect each model update to its retained supervision.

\section{Prompt Templates}
\label{sec:supp_prompts}
The following templates give the complete textual instructions used by the question generator, learner answerer, QA-pool evaluator, and benchmark scoring evaluator. Braced fields denote values substituted for the current call. The sampled frames are supplied through each model's native multimodal image interface alongside the text. Perception questions are constructed by the fixed rules in Section~\ref{sec:supp_generation}-A.

\subsection{Recognition and Understanding Question Generation}
In the ordinary template, \texttt{questions\_per\_type} is one and \texttt{total\_items} is four. The type list contains the four types for the selected dimension. The \texttt{guidance} field is formed by joining the corresponding type instructions listed immediately below the template.
\begin{lstlisting}
You are the question-generation component of CurEvo.
Inspect all sampled frames from the same video and generate {questions_per_type} distinct, concise, answerable English question(s) for each requested type.
Use only visible evidence. Do not mention frames, annotations, datasets, models, or missing context.
Do not answer the questions. Avoid yes/no questions unless explicitly required below.

Dimension: {dimension}
question_types: {requested_types_json}
Type requirements:
{guidance}

Return exactly one JSON object. The questions list must contain {total_items} items in total:
{"questions": [{"type": "one requested type", "question": "one question ending with ?"}]}
\end{lstlisting}

The recognition guidance is:
\begin{lstlisting}
- "object": Ask about a clearly visible object or what a person is holding.
- "action": Ask about the main visible action.
- "attribute": Ask about a visible color, shape, size, or appearance attribute.
- "scene": Ask where the event takes place or what scene is visible.
\end{lstlisting}
The understanding guidance is:
\begin{lstlisting}
- "temporal": Ask about event order, before/after, or temporal progression.
- "spatial": Ask about a visible spatial relation between entities.
- "causal": Ask for a cause that can be inferred from the shown evidence.
- "intention": Ask about a person's goal or intention supported by the video.
\end{lstlisting}

For NExT-QA-style generation, the following instruction is inserted after the question-generation rules:
\begin{lstlisting}
Every item must include between 3 and 5 short, distinct answer options in a "choices" array.
Exactly one option must be supported by the visible video, and the others must be plausible
but unsupported distractors. Do not state which option is correct.
\end{lstlisting}
The corresponding output schema includes the option list:
\begin{lstlisting}
{"questions": [{"type": "one requested type", "question": "one question ending with ?", "choices": ["option 1", ..., "option 5"]}]}
\end{lstlisting}
The ellipsis indicates the variable-length list in the prompt schema; an actual generated item supplies three to five concrete strings. The subsequent parser validates and shuffles these strings before answering.

\subsection{Learner Answering}
The common answer template is:
\begin{lstlisting}
Answer the video question using only the visible evidence.
{answer_rule}
Do not explain your reasoning and do not add a preamble.

Question: {question}{choice_text}
Answer:
\end{lstlisting}
For an open-ended question, the rule is:
\begin{lstlisting}
Answer with the shortest complete factual phrase supported by the video.
\end{lstlisting}
For a binary-form question, it is:
\begin{lstlisting}
Answer with exactly one word: yes or no.
\end{lstlisting}
Binary form is recognized from the leading auxiliary: is, are, was, were, does, do, did, has, have, had, can, could, will, would, or should. For a choice-based question, the rule and appended option list are:
\begin{lstlisting}
Answer exactly as 'LETTER. option text'.

Options:
A. {choice_1}
B. {choice_2}
C. {choice_3}
...
\end{lstlisting}
Only the supplied options are listed, up to E. Their order is the seeded shuffled order recorded with the candidate. The same concise answer interface is used for benchmark prediction.

\newpage
\subsection{Question Assessment}
The question evaluator uses the sampled video and the following prompt:
\begin{lstlisting}
You are the frozen evaluator in CurEvo's question assessment stage.
Inspect all video frames and evaluate the candidate question before any answer is generated.
Judge whether it is grounded in visible evidence, belongs to the requested cognitive type, and is answerable from the video.
Scores must be numbers from 0 to 1. Do not answer the question.

Dimension: {dimension}
Question type: {question_type}
Candidate question: {question}

Return exactly one JSON object:
{
  "semantic_alignment": 0.0,
  "visual_grounding": 0.0,
  "answerability": 0.0,
  "reason": "brief reason"
}
\end{lstlisting}
The numeric values in the schema indicate the expected JSON number type. The returned assessments are sanitized and combined with $H_q$ as specified in Section~\ref{sec:supp_eval}-A.

\subsection{Answer Assessment}
The answer evaluator receives the learner's prediction and the same sampled view:
\begin{lstlisting}
You are the frozen evaluator in CurEvo's answer evaluation stage.
First infer the best answer independently from the video and question. Then compare it with MODEL_ANSWER.
Do not trust or copy MODEL_ANSWER. Scores must be numbers from 0 to 1 and grounded in the shown frames.
{independent_answer_rule}

Dimension: {dimension}
Question type: {question_type}
Question: {question}{choice_text}
MODEL_ANSWER: {model_answer}

Return exactly one JSON object:
{
  "independent_answer": "your concise answer",
  "correctness": 0.0,
  "relevance": 0.0,
  "completeness": 0.0,
  "visual_consistency": 0.0,
  "reason": "brief reason"
}
\end{lstlisting}
The open-ended independent-answer rule is:
\begin{lstlisting}
The independent_answer must be a concise factual answer.
\end{lstlisting}
For choice-based questions, it is:
\begin{lstlisting}
The independent_answer must be exactly the best option as 'LETTER. option text'.
\end{lstlisting}
The optional choice text uses the same A--E layout as the learner prompt. This assessment produces the answer-quality fields and the recorded independent answer. The accepted learner answer supplies the recognition or understanding training target in the paper configuration.

\subsection{Benchmark Semantic Scoring}
After a benchmark prediction is generated, the scoring evaluator receives the video and this template:
\begin{lstlisting}
Benchmark semantic evaluation.
Judge whether PREDICTION is semantically equivalent to REFERENCE for the question. Ignore harmless wording differences.

Use this numeric rubric exactly:
5 = fully correct and semantically equivalent;
4 = essentially correct with a minor omission;
3 = partially correct but incomplete;
2 = weakly related but substantially wrong;
1 = incorrect or unrelated.
Set equivalent=true only for scores 4 or 5, and equivalent=false for scores 1 to 3.
The score must be a JSON number, not a quoted string.

Question: {question}
REFERENCE: {reference}
PREDICTION: {prediction}

Return exactly one JSON object:
{"score": 1.0, "equivalent": false, "reason": "brief reason"}
\end{lstlisting}
The schema illustrates the expected types. The numeric rubric controls the returned score, and the strict-match and binary shortcuts described in Section~\ref{sec:supp_protocol}-B are applied before a semantic-scoring call.

\subsection{JSON Repair Instruction}
When a structured response cannot be parsed, the original task prompt is repeated with the following appended instruction:
\begin{lstlisting}
Your previous output was invalid. Return one valid JSON object only, without Markdown.
\end{lstlisting}
The interface makes at most two such retries. The generation and assessment prompts retain their original semantic instructions during repair.

\section{Worked Examples}
\label{sec:supp_examples}
The examples in this section are constructed calculations that illustrate the implemented rules. Their sequences and scores are explanatory inputs, rather than additional experimental measurements.

\subsection{Controlled View and Canonical Targets}
Consider eight sampled frames denoted by their original positions $[0,1,2,3,4,5,6,7]$. The counting question has target 8. Marking zero-based position 1 gives the localization target ``second,''\ which is canonicalized to 2.

Swapping positions 1 and 6 produces
\begin{equation*}
[0,6,2,3,4,5,1,7].
\end{equation*}
The disorder question receives target ``no.''\ Replacing the frame at position 2 with its predecessor instead produces
\begin{equation*}
[0,1,1,3,4,5,6,7],
\end{equation*}
which gives target ``no''\ for consistency. An unchanged sequence gives ``yes''\ for the corresponding order or consistency question. Each label follows the saved transformation.

Table~\ref{tab:supp_canonical} illustrates how different answer surfaces map to the training interface. Canonicalization follows successful assessment and preserves the original target in metadata.

\begin{table}[t]
\caption{Constructed answer-normalization examples. Option letters refer to the shuffled choice order.}
\label{tab:supp_canonical}
\centering
\footnotesize
\setlength{\tabcolsep}{3pt}
\renewcommand{\arraystretch}{1.1}
\begin{tabularx}{\columnwidth}{p{0.24\columnwidth}Y>{\raggedright\arraybackslash}p{0.24\columnwidth}}
\toprule
\textbf{Type} & \textbf{Answer surface} & \textbf{Target}\\
\midrule
Counting & There are eight. & \texttt{8}\\
Localization & The second frame. & \texttt{2}\\
Disorder & No. & \texttt{no}\\
Object & A red kettle. & \texttt{red kettle}\\
Choice & C. opening a door & \texttt{C}\\
\bottomrule
\end{tabularx}
\end{table}

\subsection{Question Gate Calculation}
The counting question contains nine words, ends in exactly one question mark, starts with ``how,''\ and contains none of the clarity markers. Therefore, all four heuristic components equal one and $H_q=1$.

For an illustrative evaluator output with semantic alignment 0.90, visual grounding 0.95, and answerability 0.85, the semantic mean is $S_q=0.90$. The final score is
\begin{equation*}
Q=0.35(1)+0.65(0.90)=0.935.
\end{equation*}
It passes the initial question threshold 0.62 and proceeds to learner answering. If all three sanitized semantic fields instead equal zero, the configured fallback substitutes $H_q$ for each field, records its use, and yields $Q=1$.

\subsection{Deterministic Answer Gate}
For the same eight-frame counting question, the answer ``eight''\ parses to 8. Its type, conciseness, and nonempty indicators are all one, so $H_a=1$. A match with the pseudo label gives $(m,v,m,m)=(1,1,1,1)$, hence $S_a=1$ and $A=1$. The accepted target is the canonical digit 8.

For the constructed answer ``seven,''\ the format remains valid and $H_a=1$, while $m=0$ and $v=1$. Thus,
\begin{equation*}
S_a=0.25,\qquad
A=0.35(1)+0.65(0.25)=0.5125.
\end{equation*}
The pseudo-label mismatch determines rejection, including when an adaptive numerical threshold is 0.50. The record still contributes correctness zero to the current-round competence statistic. This example separates numerical quality aggregation from the deterministic admission requirement.

\subsection{Curriculum Ratio Update}
Suppose the current ratios are the initial $(0.50,0.30,0.20)$, and a constructed round yields competence $(0.80,0.60,0.40)$. With sensitivities $(0.50,0.75,1.00)$, Eq.~\eqref{eq:supp_ratio} gives
\begin{equation*}
\begin{aligned}
u_{\mathrm{perception}}&=0.50[1+0.50(0.20)]=0.55,\\
u_{\mathrm{recognition}}&=0.30[1+0.75(0.40)]=0.39,\\
u_{\mathrm{understanding}}&=0.20[1+1.00(0.60)]=0.32.
\end{aligned}
\end{equation*}
Their sum is 1.26, and normalization gives approximately $(0.4365,0.3095,0.2540)$. All intermediate shares exceed the floor, so the final normalization preserves these values. Allocating 90 videos by largest remainders gives 39 perception, 28 recognition, and 23 understanding assignments. The total remains 90, while the lower-competence dimensions receive increased relative coverage.

\subsection{Adaptive Threshold Calculation}
For a threshold associated with competence 0.40, the quantile probability is
\begin{equation*}
p=0.45+0.30(0.40)=0.57.
\end{equation*}
Consider five illustrative scores:
\begin{equation*}
[0.50,0.60,0.70,0.80,0.90].
\end{equation*}
Linear interpolation uses index $(5-1)0.57=2.28$, giving
\begin{equation*}
\widehat{\tau}=0.70+0.28(0.80-0.70)=0.728.
\end{equation*}
Starting from 0.62, the smoothed threshold is
\begin{equation*}
\tau_{\mathrm{next}}=0.70(0.62)+0.30(0.728)=0.6524.
\end{equation*}
This lies within $[0.50,0.90]$. It governs the next round; the records used in this calculation retain the admission decisions made with the preceding threshold.

\subsection{Informativeness and Training Mass}
For the illustrative target ``opening wooden door,''\ the token sequence has three distinct tokens, all of which are content tokens. Two tokens have length at least five. Therefore,
\begin{equation*}
U_{\mathrm{len}}=3/12,\quad
U_{\mathrm{div}}=1,\quad
U_{\mathrm{spec}}=2/3,
\end{equation*}
and
\begin{equation*}
U=0.35(0.25)+0.35(1)+0.30(2/3)=0.6375.
\end{equation*}
With dimension ratio 0.30 and answer score 0.90, its raw priority is $0.30\times0.90\times0.6375=0.172125$. Normalization over the full retained pool converts this priority into its training weight.

For a separate choice-balancing illustration, suppose the choice subset carries total mass 0.40, with masses 0.30 and 0.10 on the two represented letters. Eq.~\eqref{eq:supp_choice_balance} multiplies their record weights by $2/3$ and 2, respectively. Each letter then carries mass 0.20, preserving the subset's total mass. The effective factor $Nw_i'$ subsequently determines the record's contribution after its target-token losses are averaged.

\end{document}